\documentclass[times, 5p, number, twocolumn, sort&compress]{elsarticle}
\usepackage[utf8]{inputenx}
\usepackage[T1]{fontenc}
\usepackage{graphicx}
\usepackage[pdftex,
             unicode = true,
             bookmarks = true,
             bookmarksnumbered =b false,
             bookmarksopen = false,
             breaklinks = false,
             backref = false,
             colorlinks = true,
             citecolor=black,        
             filecolor=black,      
             urlcolor=black,           
            linkcolor=black          
           ]{hyperref}
\usepackage{hypernat}
\usepackage{xspace}

\usepackage{amsfonts}
\usepackage{amssymb}
\usepackage{marvosym}
\usepackage{textcomp}
\usepackage{pifont}
\usepackage{booktabs}
\usepackage{subfig}
\usepackage{array}
\usepackage[version=3]{mhchem}
\usepackage{xspace}
\usepackage{url}
\makeatletter
\def\url@cbstyle{%
  \@ifundefined{selectfont}{\def\UrlFont{}}{\def\UrlFont{\small}}}
\makeatother
\usepackage{tikz}
\usepackage{siunitx}
\colorlet{iphtblue}{black}
\usetikzlibrary{decorations.pathreplacing}
\usetikzlibrary{matrix} 
\usetikzlibrary{patterns}
\usepackage{fixltx2e}

\providecommand{\processdelayedfloats}{}

\DeclareMathOperator{\Sens}{Sens}
\DeclareMathOperator{\Spez}{Spec}
\DeclareMathOperator{\PPV}{PPV}
\DeclareMathOperator{\NPV}{NPV}

\DeclareMathOperator{\Z}{\mathbf{Z}}
\DeclareMathOperator{\Zf}{Z}
\DeclareMathOperator{\D}{\mathbf{\Delta}}

\DeclareMathOperator{\PRmat}{\mathbf{G}}

\DeclareMathOperator{\MAE}{MAE}
\DeclareMathOperator{\RMSE}{RMSE}
\DeclareMathOperator{\wMAE}{wMAE}
\DeclareMathOperator{\wRMSE}{wRMSE}
\DeclareMathOperator{\wRMAE}{wRMAE}
\newcommand{\AND}{AND\xspace}

\newcommand{\nkl}{\ensuremath{{n_g}}\xspace}

\newcommand{\Kl}{\ensuremath{G}\xspace}
\newcommand{\Klref}{\ensuremath{R}\xspace}
\newcommand{\Klpred}{\ensuremath{P}\xspace}
\newcommand{\kl}{\ensuremath{g}\xspace} 
\newcommand{\klref}{\ensuremath{r}\xspace}
\newcommand{\klpred}{\ensuremath{p}\xspace}

\newcommand{\grade}{\ensuremath{^\circ}}
\newcommand{\ie}{\textit{i.\,e.}\xspace}

\newcommand{\eg}{\textit{e.\,g.}\xspace}
\newcommand{\Eg}{\textit{E.\,g.}\xspace}
\newcommand{\mum}[1][\,]{#1\textmu m\xspace}

\renewcommand{\th}[1][th]{\textsuperscript{#1}\xspace}

\newcommand{\Gdl}{^\text{weak}}
\newcommand{\Luk}{^\text{strong}}
\newcommand{\mae}{^{\MAE}}
\newcommand{\rmse}{^{\RMSE}}
\newcommand{\prd}{^\text{prod}}
\newcommand{\opt}{^\text{opt}}
\newcommand{\pes}{^\text{pess}}
\newlength{\hlw}
\newcommand{\fancybox}[3][red]{
\begin{tikzpicture}
\node [draw=#1, very thick, rectangle, rounded corners, inner sep=10pt](box){%
  \begin{minipage}{\linewidth-20pt}
    \vspace{0.5\baselineskip}
    #3
  \end{minipage}
};
\node[fill = #1, text = white, right=5mm, rounded corners] at
(box.north west) {\sffamily\bfseries\large #2};
\end{tikzpicture}%
} 

\begin{document}
\begin{frontmatter}
\title{Validation of Soft Classification Models using Partial Class Memberships: An Extended Concept of Sensitivity \& Co. applied to the Grading of Astrocytoma Tissues}
\author[units,ipht]{Claudia~Beleites\corref{cor}}
\ead{Claudia.Beleites@ipht-jena.de}
\cortext[cor]{Corresponding author}
\author[tud]{Reiner~Salzer}
\ead{reiner.salzer@chemie.tu-dresden.de}
\author[units]{Valter~Sergo}
\ead{sergo@units.it}
\address[units]{Center of Excellence for Nanostructured Materials and Department of Engineering and Architecture, University of Trieste, Italy}
\address[ipht]{Department of Spectroscopy and Imaging, Institute of Photonic Technology, Jena, Germany}
\address[tud]{Department of Chemistry and Food Chemistry, Dresden University of Technology, Dresden/Germany}

\begin{abstract}
  We use partial class memberships in soft classification to model uncertain labelling and mixtures
  of classes. Partial class memberships are not restricted to predictions, but may also occur in
  reference labels (ground truth, gold standard diagnosis) for training and validation data.

  Classifier performance is usually expressed as fractions of the confusion matrix, such as sensitivity,
  specificity, negative and positive predictive values.  We extend this concept to soft
  classification and discuss the bias and variance properties of the extended performance
  measures. Ambiguity in reference labels translates to differences between best-case, expected and
  worst-case performance. We show a second set of measures comparing expected and ideal performance
  which is closely related to regression performance, namely the root mean squared error $\RMSE$ and
  the mean absolute error $\MAE$.
  
  All calculations apply to classical crisp as well as to soft classification (partial class
  memberships as well as one-class classifiers). The proposed performance measures allow to test
  classifiers with actual borderline cases. In addition, hardening of e.g. posterior probabilities
  into class labels is not necessary, avoiding the corresponding information loss and increase in
  variance.

  We implemented the proposed performance measures in R package ``softclassval'' which is available
  from CRAN and at \url{http://softclassval.r-forge.r-project.org}.

  Our reasoning as well as the importance of partial memberships for chemometric classification is
  illustrated by a real-word application: astrocytoma brain tumor tissue grading (80 patients,
  37\,000 spectra) for finding surgical excision borders.  As borderline cases are the actual target
  of the analytical technique, samples which are diagnosed to be borderline cases must be included in
  the validation.
\end{abstract}
\end{frontmatter}

\noindent\fancybox[blue!50!black]{Accepted Author Manuscript}{NOTICE: this is the author's version of
  a work that was accepted for publication in Chemometrics and Intelligent Laboratory
  Systems. Changes resulting from the publishing process, such as peer review, editing, corrections,
  structural formatting, and other quality control mechanisms may not be reflected in this
  document. Changes may have been made to this work since it was submitted for publication. A
  definitive version was subsequently published in Chemometrics and Intelligent Laboratory Systems,
  122 (2013), 12 -- 22,
  \href{DOI: 10.1016/j.chemolab.2012.12.003}{http://dx.doi.org/10.1016/j.chemolab.2012.12.003}.\\
  The manuscript is also available at \href{arXiv no. 1301.0264}{http://arxiv.org/abs/1301.0264},
  including the supplementary figures and tables.  }


\section*{Notation}
Throughout this paper, we use the following symbols:                                                         \\
\begin{footnotesize}
\begin{tabular}{>{$}l<{$}l}
  \toprule
  \text{Symbol}                                 & Meaning\hfill                                              \\
  \cmidrule(lr){1-1}\cmidrule(lr){2-2}
  n, \nkl                                       & number of samples and classes, respectively                \\
  \Kl \in \{1, \ldots, \nkl \}                  & crisp class (label)                                         \\
  \kl \in [0, 1]^\nkl                           & class membership (row) vector                              \\
  \PRmat^{(n \times \nkl)}                      & matrix of class memberships                                \\
  \klpred, \klref~\text{instead of}~ \kl & distinguish prediction and reference                        \\
  \Z^{(\nkl\ \mathrm{ref.}\ \times\ \nkl\ \mathrm{pred.})}                       & confusion matrix for $n$ samples                                           \\
\Zf (\klpred, \klref)  & function to calculate elements of $\Z$ for single samples \\
  \D^{(\nkl\ \mathrm{ref.}\ \times\ \nkl\ \mathrm{pred.})}                       & residual confusion matrix                                  \\
  \Sens^{operator}_\Kl                         & sensitivity wrt. class \Kl, calculated using \emph{operator}) \\
  \wedge                                        & conjunction (\AND-operator)                                \\
  \neg                                          & negation (NOT-operator)                                    \\
  \bottomrule
\end{tabular}
\end{footnotesize}                                                                                   

\section{Introduction}
\label{sec:introduction}

Validation of chemometric models is a crucial step: it is not enough to train a good model, but its
quality actually needs to be demonstrated with representative test samples. Thus, one firstly needs a plan
for obtaining suitable test samples and decide whether \eg cross validation is appropriate, or
whether unknown future samples are needed. An excellent discussion of such considerations is given by
Esbensen and Geladi \cite{Esbensen2010}. Secondly, the performance on the basis of the results is
described by suitable quantitative measures, such as the root mean squared error (RMSE) in
calibration or sensitivity, specificity and the like for classifiers. 

This paper focusses on the second aspect, although the motivation for this study did arise from the
first requirement. In the following section, we introduce a tumor tissue grading application where
representative test sets need to include ambiguous samples, \ie samples that according to the
reference labelling (ground truth, gold standard diagnosis) partially belong to more than one
class. We then show how to extend well-known classifier performance measures to work with partial
class memberships in the reference labels. We next apply these to our tumor classifier, and finally
discuss some more properties of the extended performance measures when applied to test samples that
are unambiguously assigned to their class by the reference labels.

\subsection{Application: Grading of Astrocytoma Tissues}
We illustrate the use of partial memberships with a bio-spectroscopic three class classification
problem. A detailed description of the application, including experimental details and
spectroscopic interpretation, has already been published  \cite{Beleites2011}.
Briefly, gliomas are the most common primary brain tumors. Among them, the astrocytomas are the
largest subgroup. The world health organization (WHO) distinguishes four grades of astrocytomas
according to their histology and clinical behavior \cite{VandenBerg1992, Louis2007, Kros2007}.
Astrocytomas \grade II tend to further de-differentiate and gain in malignancy.  Astrocytomas \grade
III are malignant, and glioblastomas (\grade IV; GBM) are the most undifferentiated
gliomas. Astrocytomas \grade III and GBM can originate from lower grade
tumors, or appear \emph{de novo} \cite{VandenBerg1992, Duran2007}. Pilocytic astrocytoma (\grade
I) are predominantly juvenile and clinically distinct tumors and are \emph{not} considered here.

Glioma treatment includes surgical excision, if possible. The complete removal of the tumor is one
of the most important factors for the prediction of the recurrence-free survival time of the patient
\cite{Duran2007, Stupp2007}.  Böker \cite{Moskopp2004} reports that complete removal under surgical
microscope reduces the number of tumour cells by 90\,--\,95\,\%, still leaving about
10\textsuperscript{10} tumour cells in the patient's brain.  Tumor surgery outside the brain often
applies ample safety margins around the tumor to ensure that all tumour cells are removed. This is
not possible in brain surgery as the normal brain tissue \emph{must} be preserved. An additional
difficulty arises from the infiltrative growth of the gliomas: the tumor border is hardly
visible. Within 2\,cm distance from the solid tumour, still about 10\,\% of the cells are tumour
cells and even more than 4\,cm outside the solid tumor glioma cells are
found\cite{Moskopp2004}. Thus, although complete removal of the tumor is desired, the surgeon
often decides to remove only the malignant part of the tumor. Stereo-navigation based on
pre-operative imaging such as magnetic resonance tomography (MRT) is used routinely to delineate the
excision border, but the precision is limited by the brain shift during surgery.  This constitutes
the need for additional tools that help surgeons in finding the proper excision border
\textit{in-situ} and \textit{in-vivo}.

The WHO grading scheme lists (morphological) properties of tissues. Conceptually, it is a
traditional classification system in the sense that a set of classes is defined, and each sample
belongs to exactly one of the classes. The tumor-biological reality, however, is not as distinct as
the WHO grading scheme and changes at the molecular level do not necessarily occur at the same time
as the changes in the morphology diagnosed during histological grading of the tumors
\cite{Schwartzbaum2006, Marko2011}. Neuropathologists frequently spot areas where cells are actually
in the process of de-differentiation, i.e. in the transition from one class to the next.  This leads
to ambiguity in the description of those areas.

Another type of ambiguous diagnosis states that a tissue consists of a mixture of cells of different
grades, \eg tumor cells infiltrating normal tissue.  This ambiguity can occur if the measurements
spatially do not resolve cells.  Diagnosis at single cell level, however, is not practical for
intra-surgical guidance. The working precision of the surgeons (up to ca.\ 1\,mm) requires
corresponding spatial resolution of the diagnostic tool: too high spatial resolution not only means
longer measurement times and/or undersampling but also confronts the surgeon with too detailed
information in a time-critical situation.

Both types of ambiguity occur in our example application, grading of brain tumor tissue for
\emph{intra-surgical} decision.  Note however, that grading of the actually measured tissue is
different and easier than grading of the patient's tumor. A detailed discussion of the
differences between these two distinct grading tasks has been given in \cite{Beleites2011}.

\subsection[Crisp and Soft Classification]{Crisp and Soft Classification}
\label{sec:crisp-soft-class}

\emph{Crisp} classification requires each sample to belong to exactly one of the $n_g$
pre-defined classes.  This restriction can be relaxed in two independent ways.
\begin{description}
\item[Multiple membership:] A sample may belong to \emph{more than} one class (or no class at
  all). 
\end{description}

\begin{description}
\item[Partial membership:] A sample may belong \emph{partially} to any given class. 
\end{description}

Multiple membership is often associated with one-class classifiers. One-class classifiers model each
class independently of the other classes \cite{Brereton2011,
  Brereton2009, Tax2001}. While this is not the case in our application (the tissue classes are
mutually exclusive) the reasoning presented here works for one-class classifiers as well. We refer to
the boundary condition of crisp classifiers that each sample must belong to exactly one class as
``closed world''. ``Open world'' intermediate results can be transformed in closed world results by
``winner takes all'' or soft max (for partial memberships) rules.

The second concept is in analogy to the transition from hard (crisp) cluster analysis to fuzzy
cluster analysis: the degree of belonging to a class is represented by a continuous membership
value. In the remote sensing community, the term \emph{soft classification} has already been
established for such partial class memberships \cite{Foody2002, Binaghi1999, Pontius2006,
  SilvanCardenas2008, Chen2010}, so we adapt this terminology. Also, \emph{fuzzy} usually refers to
ambiguity as opposed to uncertainty, but the partial memberships can denote both. Partial class
memberships can be treated as intermediate results which are then ``hardened'' into crisp class
memberships.

In chemometric \emph{modelling}, the term ``soft'' is often used in different ways. Hard vs. soft
modelling can refer to the amount of prior knowledge that is reflected in the model equations. While
hard models fit equations that are derived from strong assumptions or first principles (\eg order of
reaction for kinetic studies), soft models make less assumptions and model empiric approximations
(\eg fitting some sigmoid) \cite{Wold1977, Juan2000}. The ``soft'' in Soft Independent Modelling of
Class Analogies (SIMCA) comes from this distinction. SIMCA is an established and widespread one-class
classification model, see \eg \cite{Brereton2009, Varmuza2009, Oliveri2012} that has also been used
in the context of vibrational spectroscopic diagnosis or distinction of biological tissues
\cite{Krafft2006, Khanmohammadi2009} and cells \cite{Heraud2006}.  Varmuza and Filzmoser
\cite{Varmuza2009}, however, seem to use the term ``soft'' synonymous for one-class classifiers and
Brereton \cite{Brereton2011} defines ``soft'' as allowing overlap in the (feature) space assigned to
each class.  ``Soft'' takes yet another meaning for soft margins of support vector machines (SVM)
where soft margins allow samples in between the margins of the SVM in feature space, even though they
are labelled as belonging to exactly one class \cite{Brereton2010}.

In contrast to these soft aspects of chemometric models, we use the term \emph{soft} in this paper
with respect to class labels, and contrast it to \emph{crisp}.  The performance measures we discuss
in the present paper therefore work regardless of these various soft aspects of classification
models: validation usually treats the classifier as a black box that calculates class membership from
the spectrum (feature vector) of a test sample. Our performance measures can be calculated just the
same way whether the classifier uses hard or soft modelling, models classes independently or in
distinction of the other classes, or whether it is a SVM with or without soft margin.

While partial memberships are widely used in cluster analysis (fuzzy c-means clustering is well
established for the analysis of spectra of biological tissues \cite{Lasch2004, Steller2006,
  Krafft2009, Bonifacio2010}), this is not the case for chemometric classification. Classification
addresses qualitative questions. But qualitative analysis is usually carrid out by chemometric
quantification (regression) which is then evaluated with respect to a threshold or limit. Calibration
models adequately cover chemical composition, but  are not appropriate for many
bio-spectroscopic classification problems.

We employ (row) vectors $\kl \in \{0, 1\}^\nkl$ with the elements corresponding to the sample's class
$\kl_\Kl = 1$ and all other elements 0 to express the crisp class membership of a sample, which can
be combined into a membership matrix with each row corresponding to one sample.  We will use the term
``crisp'' label or sample also for samples that \emph{happen} to have all class memberships either 0
or 1, and ``soft'' for samples where at least one membership value is not exactly 0 or 1. Thus, there
may be (and often are) crisp samples also in a soft data set.  The boundary condition for
closed-world classifiers is $\sum_{j = 1}^\nkl \kl_j = 1$. Partial memberships allow the elements of
\kl to take any value between 0 and 1: $\kl \in [0, 1]^\nkl$.

Partial memberships can arise from two different concepts:
\begin{description}
\item[probability or uncertainty:] This is common for predictions like posterior probabilities, but
  may also be the case for reference labels.
\item[mixtures of the underlying classes] as in homogeneous mixtures or heterogeneous mixtures where
  the heterogeneity is not resolved by the measurement. In chemistry, this is closely related to the
  concept of concentration and thus to calibration. Non-chemical fields frequently use fuzzy set
  theory.
\end{description}
In practice both aspects can arise for one and the same problem.  
In biomedical applications, uncertain references arise \eg from the pathologist expressing
uncertainty: ``there may be tumor cells between these normal cells'', or from
disagreement among a panel of pathologists. In our experiments, the transfer of the histological
diagnosis onto the measurement of a parallel section is an additional source of uncertainty.
Our samples also contain two different types of mixtures: firstly, a tissue may consist of cells of
different cell types, while the individual cells are not resolved by the measurement's spatial
resolution. The second type of mixture are currently de-differentiating tissues, e.g. cells
undergoing the transition from \grade II to \grade III, which are therefore between the ordered
classes.

Partial class memberships in classification may be used and discussed at three levels:
\begin{description}

\item[Soft predictions] are widely used: posterior probabilities of linear or quadratic discriminant
  analysis (LDA and QDA) or logistic regression (LR); the voting proportion of $k$ nearest neighbors
  (kNN), random forests, etc. Soft predictions are frequently considered an intermediate result and
  are then hardened by thresholds.
\item[Soft training samples] can be used by methods like LR, artificial neural networks or partial
  least squares discriminant analysis, PLS-DA), but up to now this is rarely done.
\item[Soft test samples,] \ie samples with ambiguous or uncertain reference (ground truth, gold
  standard diagnosis) are the topic of this paper.
\end{description}

For classifier training, traditionally either crisp reference labels are enforced and/or borderline
cases\footnote{Here, we use the terms borderline case and ambiguous sample synonymously and
  exclusively with respect to the true class membership. Spectra that are spectroscopically in between the typical spectra
  of classes will be referred to as ``close to the class boundary''.}  are excluded. This raises
several issues that are avoided by allowing soft labels. Hardening of a continuous variable implies
a loss of information. Dichotomization of a logistically distributed random variable deletes at
least 25\,\% of the information \cite{Fedorov2009}. In biomedical spectroscopy, crisp reference
labels are often enforced by requesting the pathologist to assign the sample to exactly one of the
classes, even if the pathologist describes the sample as currently undergoing de-differentiation or
consisting of mixed cell populations.  In practice, the pathologist refuses to diagnose certain
samples unambiguously (leading to exclusion of the sample). For other samples the written-out
diagnosis contains information about the ambiguity that is not reflected by the assigned crisp
class. The same applies to diagnoses given by a panel of pathologists. Again, either the majority
class is used (removing the information contained in the disagreement) or the case is excluded.  For
the brain tumour patients, crisp diagnoses given by the local neuropathologist and neuropathologists
from the tumour reference center often differ more than the written-out diagnoses. This is in
accordance with hardening as possible cause of further variance. Likewise, the results of the panel
diagnosis published by Kendall \emph{et al.} \cite{Kendall2003} have higher discrepancy for
intermediate classes \cite{Beleites2011}.

In any case, one either uses possibly inappropriate descriptions as reference or gold standard
diagnosis, or reduces the available number of samples. In bio-spectroscopy, where frequently hundreds
or thousands of variates are measured for tens of patients only, this is a critical issue. In our
application, $\frac{1}{3}$ of the patients and almost half of the spectra would have to be excluded.
Moreover, excluding borderline samples comes at the risk of overestimating class separation: throwing
away all difficult cases creates an easy problem. While such filtering and the corresponding output
of ``no certain prediction possible'' are appropriate for certain analytical tasks, this is not the
case in our application.
Borderline samples are actual examples of the class boundaries. Excluding them from classifier
training means excluding most valuable samples.  The more so in our application, as these are also
examples of the actual target samples of the glioma grading technique.

While one may also arrive at a good classifier with completely unambiguous training data, the
validation must use carefully collected test samples: samples representative for the field of
use. For our astrocytoma application these, again, are the borderline cases. Validation methods for
soft labeled samples are thus even more crucial than the respective training strategies.

The remote sensing community has been using soft classification for a long time to describe the
mixtures due to low spatial resolution. Proposals for validation of soft classifiers are reported in
the literature \cite{Foody2002, Binaghi1999, Pontius2006}\,--\,yet, it is still considered an
unsolved problem \cite{SilvanCardenas2008, Chen2010}. We critically discuss these proposals below.

\section{Classifier Performance: the Confusion Matrix and fractions thereof}
For convenience, we abbreviate sums of particular parts of the confusion matrix $\Z$ as follows:
summation includes all possible indices according to the conditions written in the indices. \Eg $\sum
\Z_{i,\Klpred}$ stands for $\sum_{i = 1}^\nkl \Z_{i,\Klpred}$ (sum all elements of column $\Klpred$), $\sum \Z_{i
  \neq \Kl ,\Klpred}$ means $\sum_{i \in \{1, \ldots, \nkl | i \neq \Kl\}} \Z_{i ,\Klpred}$ (sum all
elements except that in row $\Kl$ of column $\Klpred$). $\sum_n$ is the sum over all samples.

\subsection[Hard Classification]{Hard Classification}
\newcommand{\zufrac}[5]{
  \matrix at (#1) (#2-zaehler) [matrix of nodes, cells = {draw}, 
  nodes = {draw, text width = 1ex, text height = 0.75ex, text depth = 0.25ex},
  execute at empty cell = {\node{};},
  row sep = -0.5pt,
  column sep = -0.5pt,
  ampersand replacement=\&,
  #3
]{#4};
\draw[very thick] ([below = 0.5ex] #2-zaehler.south west) ++(-.5ex, 0pt) node[coordinate] (#2-west) {}  -- 
                  ([below = 0.5ex] #2-zaehler.south east) -- ++(+.5ex, 0pt) node[coordinate] (#2-east) {};

 \matrix (#2-nenner) at (#2-zaehler.south)  [below = 1ex, matrix of nodes, cells = {draw}, 
  nodes = {draw, text width = 1ex, text height = 0.75ex, text depth = 0.25ex},
  execute at empty cell = {\node{};},
  row sep = -0.5pt,
  column sep = -0.5pt,
  ampersand replacement=\&
]{#5};
}
\newcommand{\tdiag}[1][0.75em]{\tikz[x=#1, y=#1, tight background]{
    \draw [ultra thin] (0, 0) rectangle (1, 1);  
    \draw (0, 0) -- (1, 1);
  }%
}
\newcommand{\tmdiag}[1][0.75em]{\tikz[x=#1, y=#1, tight background]{
    \draw [ultra thin] (0, 0) rectangle (1, 1);  
    \draw (0, 1) -- (1, 0);
  }%
}

\newcommand{\thv}[1][0.75em]{\tikz[x=#1, y=#1, tight background]{
    \draw [ultra thin] (0, 0) rectangle (1, 1);  
    \draw (0, .5) --++ (1, 0);
    \draw (.5, 0) --++ (0, 1);
  }%
}
\newcommand{\tp}[1][0.75em]{\tikz[x=#1, y=#1, tight background]{
    \draw [ultra thin] (0, 0) rectangle (1, 1);  
    \fill (0.5, .5) circle (#1/25);
  }%
}

\usetikzlibrary{backgrounds} 

\tikzstyle{thin}+=[line width=0.5pt]
\tikzstyle{shade}=[fill = gray]
\tikzstyle{eqmat}=[matrix of math nodes, below, 
  ampersand replacement=\&, column sep = 0pt,
  nodes = {anchor = base west, fill =yellow}]
 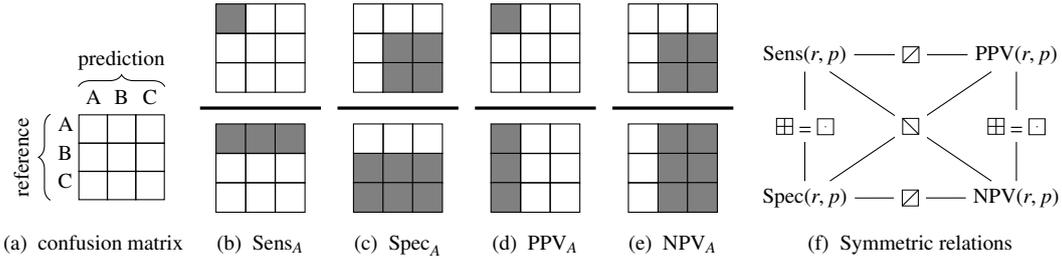
\begin{figure*}
 \centering
\subfloat[\label{fig:Z} confusion matrix]{
\begin{footnotesize}
\begin{tikzpicture}
  \matrix (cm) [matrix of nodes, 
  nodes = {text width = 1.5ex, text height = 1ex, text depth = 0.5ex, anchor = base east},
  execute at empty cell = {\node[draw]{};},
  row sep = -0.5pt,
  column sep = -0.5pt,
  ampersand replacement=\&,
  ]{
    ~ \& A \& B \& C \\
    A \&   \&   \&   \\
    B \&   \&   \&   \\
    C \&   \&   \&   \\
  };
  \draw[decorate,decoration={brace,amplitude = 1.25ex}] 
  (cm-1-2.north west) -- (cm-1-4.north east) 
  node [midway, above = 1ex, text width = 12ex, text badly centered] {prediction};
  \draw[decorate,decoration={brace, amplitude = 1.25ex}] 
  (cm-4-1.south west) -- (cm-2-1.north west) 
  node[midway, xshift = -3ex, text width = 12ex, text badly centered, rotate = 90] 
  {reference} ;
\end{tikzpicture}
\end{footnotesize}}
\subfloat[\label{fig:sens}  $\Sens_A$]{
  \begin{tikzpicture}
    \zufrac{0,0}{senscl}{}
    {
      |[shade]| \&           \&            \\
                \&           \&            \\
                \&           \&            \\
    }{
      |[shade]| \& |[shade]| \& |[shade]|  \\
                \&           \&            \\
                \&           \&            \\
    }
 \end{tikzpicture}
}
\subfloat[\label{fig:spez} $\Spez_A$]{
  \begin{tikzpicture}
    \zufrac{0,0}{senscl}{}
    {
                \&           \&            \\
                \& |[shade]| \& |[shade]|  \\
                \& |[shade]| \& |[shade]|  \\
    }{
                \&           \&            \\
      |[shade]| \& |[shade]| \& |[shade]|  \\
      |[shade]| \& |[shade]| \& |[shade]|  \\
    }
   \end{tikzpicture}
}
\subfloat[\label{fig:ppv} $\PPV_A$]{
  \begin{tikzpicture}
    \zufrac{0,0}{senscl}{}
    {
      |[shade]| \&           \&            \\
                \&           \&            \\
                \&           \&            \\
    }{
      |[shade]| \&           \&            \\
      |[shade]| \&           \&            \\
      |[shade]| \&           \&            \\
    }
 \end{tikzpicture}
}
\subfloat[\label{fig:npv} $\NPV_A$]{
  \begin{tikzpicture}
    \zufrac{0,0}{senscl}{}
    {
                \&           \&            \\
                \& |[shade]| \& |[shade]|  \\
                \& |[shade]| \& |[shade]|  \\
    }{
                \& |[shade]| \& |[shade]|  \\
                \& |[shade]| \& |[shade]|  \\
                \& |[shade]| \& |[shade]|  \\
    }
   \end{tikzpicture}
}
\subfloat[\label{fig:symmetrie} Symmetric relations]{
  \footnotesize
  \begin{tikzpicture}
    \matrix (measures) [matrix of math nodes, 
    nodes = {
    },
    execute at empty cell = {\node{};},
    row sep = {15ex,between origins},
    column sep = {22ex,between origins},
    ampersand replacement=\&,
    ]{
      \Sens (r, p)  \& \PPV  (r, p)\\
      \Spez (r, p) \& \NPV  (r, p)\\
    };
      \foreach \i/\j in {1/2, 2/1}{ 
        \draw (measures-1-\i) -- (measures-2-\i) node [midway, fill = white] {\thv{} = \tp}  ;
        \draw (measures-\i-1) -- (measures-\i-2) node [midway, fill = white] {\tdiag}  ;
      }
      \draw (measures-1-1) -- (measures-2-2);
      \draw (measures-2-1) -- (measures-1-2) node [midway, fill = white] {\tmdiag}  ;
  \end{tikzpicture}
}
\label{fig:kenngr} 
\caption[]{ Confusion matrix \protect\subref{fig:Z} and characteristic fractions for sum constrained
  multi class classifiers \protect\subref{fig:sens} -- \protect\subref{fig:npv}. The parts of the
  confusion matrix summed as numerator and denominator for the respective fraction with respect to
  class $A$ are shaded. \protect\subref{fig:symmetrie} Symmetry between the measures.  \thv[1.5ex] =
  \tp[1.5ex]: mirror horizontally and vertically or at point $(\klref; \klpred) \mapsto (1-\klref;
  1-\klpred)$, \tdiag[1.5ex]: mirror at major diagonal $(\klref; \klpred) \mapsto (\klpred; \klref)$,
  and \tmdiag[1.5ex]: mirror at minor diagonal $(\klref; \klpred) \mapsto (1 - \klpred; 1 - \klref)
  $.  All symmetry elements with respect to the center of the value space at (0.5; 0.5). The icons
  use Cartesian coordinates. \subref{fig:Z} -- \subref{fig:npv} are reprinted from
  \cite{Beleites2013}, with permission from Elsevier.}
\end{figure*}

The validation results of a crisp classifier are usually tabulated in the
 confusion matrix $\Z$.  This matrix (fig.~\ref{fig:Z}) counts how many samples that truly
belong to each class (rows) were predicted to belong to that class (columns). In other words, a
sample belonging to class \Klref and predicted to belong to class \Klpred is counted in $\Z_{\Klref,
  \Klpred}$. Sometimes, a notation as function of prediction and reference is more convenient:
\begin{equation}
     \label{eq:def-crisp-confmat}
     \Z_{i, j}  = \sum_n \Zf (\klref_i, \klpred_j) 
      = \sum_n \klref_i \wedge \klpred_j.
\end{equation} 
where $\wedge$ stands for the \AND\ operator which returns 1 if and only if both reference class
membership $\klref_i$ and prediction class membership $\klpred_j$ are 1, otherwise the return value
is 0.  In order not to clutter up the notation, we indicate the sum over all samples by $\sum_n$
without introducing an index for the sample. The symbol $\Z$ for the confusion matrix will imply that
this sum is already taken, whereas $\Zf (\klref_i, \klpred_j)$ is evaluated for each sample. The
results are then summed up to give $\Z_{i, j}$.

Confusion matrices are frequently pooled (\eg $k$ confusion matrices obtained during one iteration of
$k$-fold cross validation are fused into one by matrix addition).
Confusion matrices yield a very detailed overview of a classifier's performance. Frequently, the
confusion matrix is further summarized by proportions calculated thereof. These proportions
(fig.~\ref{fig:sens} -- \ref{fig:npv}) answer questions with regard to the predictive
abilities of a classifier. Different disciplines refer to these fractions differently.  We use the
medical terminology \cite{Ellison2005, Forthofer}:
\begin{description}
\item[Sensitivity $\Sens_\Kl$:] How well does the classifier recognize samples of class \Kl?
\item[Specificity $\Spez_\Kl$:] How well does the classifier recognize that a sample does \emph{not}
  belong to class \Kl?
\item[Positive Predictive Value $\PPV_\Kl$:] Given the prediction is class \Kl, what is the probability
  that the sample truly belongs to \Kl?
\item[Negative Predictive Value $\NPV_\Kl$:] Given a prediction ``does not belong to class \Kl'', what is
  the probability that the sample truly does not belong to \Kl?
\end{description}
Note that the predictive values are the ``inverse'' (as in inverse calibration) of sensitivity and
specificity: sensitivity and specificity report the distribution of test outcomes as function of the
true disease status. In contrast, the predictive values give the distribution of true disease status
as function of the observed test outcome.

For users of the classifier, the predictive values are usually of more interest than sensitivity and
specificity: patients and doctors want to know whether \emph{this particular} patient is ill rather than whether the
test can recognize ill people; manufacturers want to know whether a product can be sold rather than
whether bad batches can be found.  Answering these questions needs to take into account the prior
probabilities of the classes (in medical diagnosis: prevalence). The relative frequencies of the
classes in the test set (row sums of the confusion matrix $\Z$) do \emph{not} necessarily reflect the
prior probabilities. Moreover, the prior probabilities can vary greatly among different populations
(consider \eg HIV tests for blood donors and drug addicts, respectively). The reported
predictive values should therefore be corrected for the different composition of test set and target
population and also specify the (assumed) composition of the target population. The same caution
applies for all measures that combine different reference classes, such as overall accuracy, the
chance agreement needed to calculate corrected performance values like the $\kappa$ statistic, etc.

Usually, these performance measures relate to medical decisions whether a certain disease is present
or absent. This corresponds to one-class classifiers. The questions translate to the following
expressions:
\begin{align}
  \Sens_\Kl & = \frac{\Z_{\Kl,\Kl}}{\sum_n \klref_{\Kl}}              \\
  \PPV_\Kl  & = \frac{\Z_{\Kl,\Kl}}{\sum_n \klpred_{\Kl}}      \\
  \Spez_\Kl & = \frac{\Z_{\neg\Kl, \neg\Kl}}{\sum_n \klref_{\neg\Kl}} \\
  \NPV_\Kl  & = \frac{\Z_{\neg\Kl, \neg\Kl}}{\sum_n \klpred_{\neg\Kl}} 
\end{align}
The membership $\kl_{\neg\Kl}$ of the dummy class ``$\neg \Kl$'' (``not class $\Kl$'') is obtained as
$\kl_{\neg\Kl} = 1 - \kl_\Kl$. The same expressions can be used for closed-world classifiers, where
the constraint in addition allows the alternative calculation as the sum of memberships to the other
classes $\kl_{\neg\Kl} = 1 - \kl_\Kl = \sum_{\kl \neq \Kl} \kl_\kl$, see fig.~\ref{fig:sens} --
\ref{fig:npv}.

In analogy to the addition of confusion matrices, the overall (multi-sample) performance is the
average of the single sample performances weighted by the denominator variable.

Like the confusion matrix, also the performance measures can be written as function, and due to the
symmetry between the performance measures (compare fig.~\ref{fig:symmetrie}), all operators can be
expressed using one basic underlying function.  Performance measures as well as the class memberships
refer to each class independently of the other classes for both one-class and closed-world
classifiers. We drop the class index for convenience:
\begin{align}
  \label{eq:sens} \Sens (\klref,\klpred)  & = \frac{\sum_n \Zf (\klref, \klpred)}{\sum_n \klref} \\
  \label{eq:spez} \Spez (\klref,\klpred)  & = \Sens (1 - \klref, 1 - \klpred)\\
  \label{eq:PPV}  \PPV  (\klref,\klpred)  & = \Sens (\klpred, \klref)\\ 
  \label{eq:NPV}  \NPV  (\klref,\klpred)  & = \Sens (1 - \klpred, 1 - \klref)
\end{align}

\subsection{Soft Confusion Matrices}
\label{sec:soft-confmat}
To generalize the confusion matrix and performance measures for soft reference and prediction, the
Boolean \AND-operator $\wedge$ (conjunction) in the definition of the crisp confusion matrix
(eq.~\ref{eq:def-crisp-confmat}) is replaced by a suitable operator for continuous-valued input in the
range $[0, 1]$.  The three main candidates are the minimum (weak conjunction), as proposed e.\,g. by
 Łukasiewicz and Gödel, the strong conjunction $\max (x + y - 1, 0)$ (Łukasiewicz), and the product
\cite{sep-logic-manyvalued, sep-logic-fuzzy, Dubois1985}. These operators reflect the ambiguity in
the performance estimate due to the ambiguity expressed by the soft memberships.

\newcommand{\prodr}[1][]{
\begin{tikzpicture}[#1]

\foreach \x/\xe in {
0/0.2, 0.3/0.4, 0.6/0.7, 0.8/1, 1.4/1.5, 1.6/2, 2.1/2.2, 2.4/2.5, 2.7/2.8, 2.9/3, 3.1/3.2, 3.3/3.6, 3.8/4, 4.1/4.2, 4.3/4.4, 4.5/4.8, 4.9/5, 5.6/5.8, 5.9/6, 6.2/6.3, 6.5/6.6, 6.8/7.1, 7.3/7.5, 7.6/7.7, 7.8/7.9, 8.2/8.4, 8.6/8.9, 9/9.1, 9.2/9.4, 9.6/9.8, 9.9/10 }{
	\fill [iphtblue] (\x em, 0em)  rectangle (\xe em, 1em);
}
\draw (0em,0em) rectangle ++(10em, 1em);
\end{tikzpicture}
}
\newcommand{\prodp}[1][]{
\begin{tikzpicture}[#1]

  \foreach \x/\xe in {
0.1/0.2, 0.3/1.1, 1.3/1.7, 1.8/2.4, 2.5/3.4, 3.5/3.6, 3.8/4.4, 4.5/4.7, 4.9/5.2, 5.4/5.7, 5.8/6.9, 7/7.4, 7.5/7.6, 7.7/8.3, 8.4/9.1, 9.2/10 }{
	\fill [iphtblue] (\x em, 0em)  rectangle (\xe em, 1em);
}
\draw (0em,0em) rectangle ++(10em, 1em);
\end{tikzpicture}
}

\newcommand{\prodZ}[1][]{
\begin{tikzpicture}[#1]

  \foreach \x/\xe in {
0.1/0.2, 0.3/0.4, 0.6/0.7, 0.8/1, 1.4/1.5, 1.6/1.7, 1.8/2, 2.1/2.2, 2.7/2.8, 2.9/3, 3.1/3.2, 3.3/3.4, 3.5/3.6, 3.8/4, 4.1/4.2, 4.3/4.4, 4.5/4.7, 4.9/5, 5.6/5.7, 5.9/6, 6.2/6.3, 6.5/6.6, 6.8/6.9, 7/7.1, 7.3/7.4, 7.8/7.9, 8.2/8.3, 8.6/8.9, 9/9.1, 9.2/9.4, 9.6/9.8, 9.9/10 }{
	\fill [iphtblue] (\x em, 0em)  rectangle (\xe em, 1em);
}
\draw (0em,0em) rectangle ++(10em, 1em);
\end{tikzpicture}
}

\newcommand{\strongr}[1][]{
\begin{tikzpicture}[#1]

\fill [iphtblue] (0, 0)  rectangle ++(5em, 1em);
\draw (0em,0em) rectangle ++(10em, 1em);
\end{tikzpicture}
}
\newcommand{\strongp}[1][]{
\begin{tikzpicture}[#1]

\fill [iphtblue] (2em, 0em) rectangle (10em, 1em);
\draw (0em,0em) rectangle ++(10em, 1em);
\end{tikzpicture}
}
\newcommand{\strongZ}[1][]{
\begin{tikzpicture}[#1]

\fill [iphtblue] (2em, 0em) rectangle (5em, 1em);
\draw (0em,0em) rectangle ++(10em, 1em);
\end{tikzpicture}
}

\newcommand{\weakr}[1][]{
\begin{tikzpicture}[#1]

\fill [iphtblue] (0, 0)  rectangle ++(5em, 1em);
\draw (0em,0em) rectangle ++(10em, 1em);
\end{tikzpicture}
}
\newcommand{\weakp}[1][]{
\begin{tikzpicture}[#1]

  \fill [iphtblue] (0em, 0em) rectangle ++(8em, 1em);
\draw (0em,0em) rectangle ++(10em, 1em);
\end{tikzpicture}
}
\newcommand{\weakZ}[1][]{
\begin{tikzpicture}[#1]
  \fill [iphtblue] (0em, 0em) rectangle ++(5em, 1em);
\draw (0em,0em) rectangle ++(10em, 1em);
\end{tikzpicture}
}
\begin{figure*}
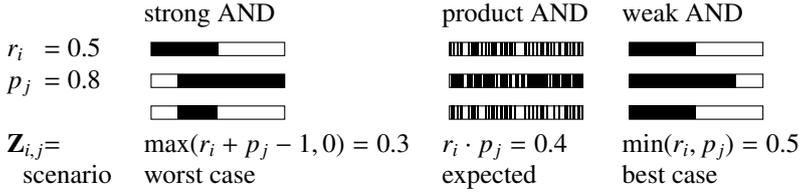

 \begin{tabular}{@{}l@{}llll}
     &              & strong AND & product AND & weak AND \\[.25ex]
    $r_i$&= 0.5    & \strongr[scale = 0.5]    & \prodr[scale = 0.5]       & \weakr[scale = 0.5]    \\
    $p_j$&= 0.8    & \strongp[scale = 0.5]    & \prodp[scale = 0.5]       & \weakp[scale = 0.5]    \\
&                   & \strongZ[scale = 0.5]    & \prodZ[scale = 0.5]       & \weakZ[scale = 0.5]    \\
    $\Z_{i, j}$ & =  & $\max (r_i + p_j - 1, 0) = 0.3$ & $r_i \cdot p_j = 0.4$            & $\min (r_i, p_j) = 0.5$       \\
\multicolumn{2}{l}{scenario}    & worst case & expected & best case 
\end{tabular}
\caption{The soft \AND-operators: hypothetical high-resolution scenarios corresponding to a low
  resolution situation with reference = 0.5 (top row) and prediction = 0.8 (middle row) membership to
  the black class. In each column, the overlap (bottom row) is obtained by the classical Boolean
  \AND: for each position, $\Z_{i, j, pos} = r_{i, pos} \wedge p_{j, pos}$, the soft conjunction is
  the fraction where both $r_i$ \AND $p_j$ belong to the black class.}
  \label{fig:explops}
\end{figure*}
The rationale behind these operators can be illustrated by a situation where low (e.\,g. spatial)
resolution causes the ambiguity (fig~\ref{fig:explops}). Say, a number of cells are in the
measurement volume of a spectrum, and half of them are cancer cells and the other half are
normal. The classifier yields a fraction of 0.8 for cancerous. In the best case, the classifier recognized
the cancerous half correctly, so the conjunction (overlap) for the ``cancer'' class is 0.5. In
the worst case, the classifier assigns ``cancer'' to normal cells. However, since at least 0.3 must
still be assigned to the correct ``cancer'' class, the overlap is 0.3. The true overlap can be
anywhere between these bounds, depending on the true distribution of cancer cells and the
distribution of cancer cells predicted by a high-resolution classifier.  E.\,g. if they are uniformly
randomly distributed, for each cell the chance that it is both cancerous and predicted to be
cancerous is 0.5 $\cdot$ 0.8 = 40\,\%, and the expected overlap is 0.4 (middle column of
fig~\ref{fig:explops}).
The top row of the supplementary figure~\ref{fig:suppl-operator} illustrates the three operators as
function of reference and predicted memberships, fig.~\ref{fig:operator-behaviour} compares them for
given reference memberships.
\begin{figure}[tb]
  \centering
  \includegraphics[width=0.6\hlw]{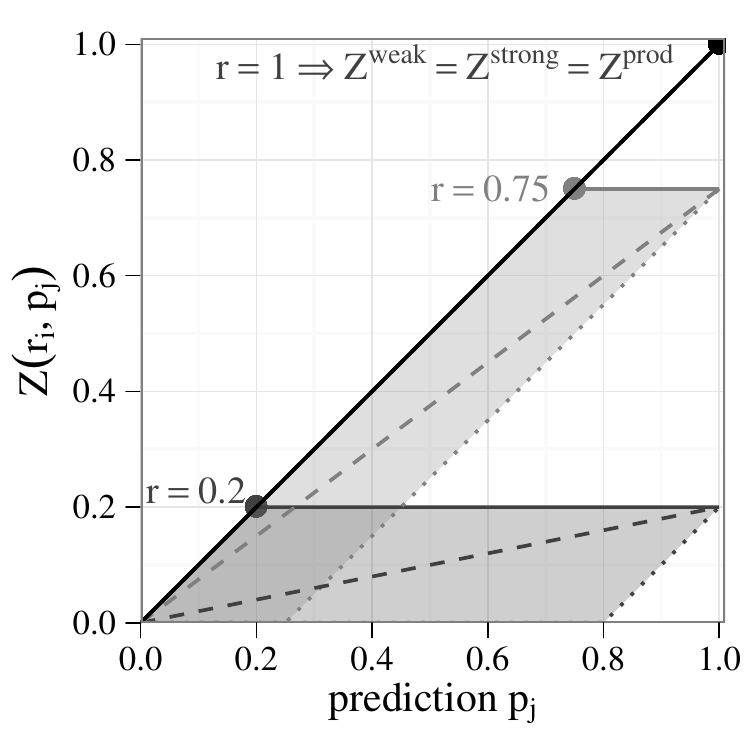}
  \caption{Behavior of the confusion matrix function $\Zf (\klref_i, \klpred_j)$ for the three
    operators for reference memberships (points) $\klref_i = $ 1 (black), 0.75 (light gray), and 0.2
    (dark gray): $\Zf\Gdl$ (upper bound of the parallelograms, continuous line), $\Zf\Luk$ (lower
    bound of the parallelograms, dotted line), and $\Zf\prd$ (dashed line). For reference class
    membership $\klref_i = $ 1 all three $\Zf$ equal the predicted membership $\klpred_j$.}
  \label{fig:operator-behaviour}
\end{figure}
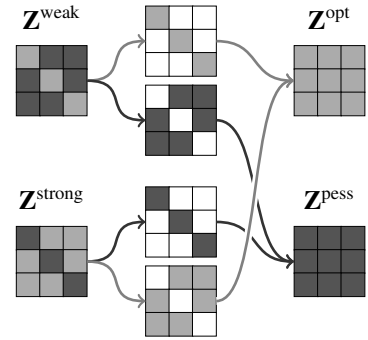
\begin{figure}[tb]
  \centering
\newcommand{\confmatod}[5]{
\matrix at (#1) (#2)  [matrix of nodes, cells = {draw, black, very thin}, 
  nodes = {draw, #3,
    text width = 1ex, text height = 0.75ex, text depth = 0.25ex, outer sep = 0pt},
  row sep = -0.1ex,
  column sep = -0.1ex,
  execute at empty cell = {\node[#4]{~};},
  ampersand replacement=\&,
  outer xsep=4ex,
  #5
]{ ~ \&   \&   \\
     \& ~ \&   \\
     \&   \& ~ \\
};
}
\begin{tikzpicture}[]
\footnotesize
\tikzstyle{sopt}=[fill=black!33]
\tikzstyle{spes}=[fill=black!67]
\tikzstyle{copt}=[black!50]
\tikzstyle{cpes}=[black!80]

\confmatod{0, 0        }{gdl}{sopt}{spes}{}
\confmatod{gdl.south}{luk}{spes}{sopt}{below = 10ex}

\confmatod{gdl.east}{gdl1}{sopt}{fill=white}{right = 5ex, above = -.33ex}
\confmatod{luk.east}{luk1}{spes}{fill=white}{right = 5ex, above = -.33ex}

\confmatod{gdl.east}{gdl2}{fill=white}{spes}{right = 5ex, below = -.33ex}
\confmatod{luk.east}{luk2}{fill=white}{sopt}{right = 5ex, below = -.33ex}

\confmatod{gdl1.east}{opt}{sopt}{}{right = 7ex, below = -.33ex}
\confmatod{luk1.east}{pes}{spes}{}{right = 7ex, below = -.33ex}
\tikzstyle{myarrow}=[->, line width=1pt]

\draw[myarrow, copt] (gdl-2-3.east) .. controls (gdl.east) and (gdl1.west) .. (gdl1-2-1.west);
\draw[myarrow, cpes] (gdl-2-3.east) .. controls (gdl.east) and (gdl2.west) .. (gdl2-2-1.west);
\draw[myarrow, cpes] (luk-2-3.east) .. controls (luk.east) and (luk1.west) .. (luk1-2-1.west);
\draw[myarrow, copt] (luk-2-3.east) .. controls (luk.east) and (luk2.west) .. (luk2-2-1.west);

\draw[myarrow, cpes] (gdl2-2-3.east) .. controls (gdl2.east) and (pes.west) .. (pes-2-1.west);
\draw[myarrow, cpes] (luk1-2-3.east) .. controls (luk1.east) and (pes.west) .. (pes-2-1.west);

\draw[line width=3pt, white, shorten >= 1pt] (luk2-2-3.east) .. controls (luk2.east) and (opt.west) .. (opt-2-1.west);
\draw[myarrow, copt] (gdl1-2-3.east) .. controls (gdl1.east) and (opt.west) .. (opt-2-1.west);
\draw[myarrow, copt] (luk2-2-3.east) .. controls (luk2.east) and (opt.west) .. (opt-2-1.west);
\normalsize
\node[above] at (gdl.north) {$\Z\Gdl$}; 
\node[above] at (luk.north) {$\Z\Luk$}; 
\node[above] at (opt.north) {$\Z\opt$}; 
\node[above] at (pes.north) {$\Z\pes$}; 
\end{tikzpicture}
\caption{Recombination of $\Z\Gdl$ and $\Z\Luk$ into $\Z\opt$ and $\Z\pes$. The diagonal of $\Z\Gdl$
  and the off-diagonal elements of $\Z\Luk$ measure the best possible performance (light gray), while
  $\Z\Luk$'s diagonal and $\Z\Gdl$'s off-diagonal elements report the worst case performance (dark
  gray). The resulting confusion matrices give the most optimistic and most pessimistic view on the
  classifier's performance in accordance with the given reference memberships and the observed
  predictions.}
  \label{fig:optpes}
\end{figure}

The \emph{weak conjunction} is the standard \AND-operator in fuzzy logic, and has been used to
compute soft confusion matrices \cite{Binaghi1999}.
\begin{equation}
  \Zf\Gdl (\klref_i, \klpred_j) = \min (\klref_i, \klpred_j)
\end{equation}
The minimum is the \emph{highest} possible overlap between prediction and reference (best case
scenario in fig.~\ref{fig:explops} and upper bound in fig.~\ref{fig:operator-behaviour}).

The \emph{strong conjunction} has been introduced for soft classifier performance by Pontius
\emph{et.al.} \cite{Pontius2006a}. It reports the \emph{lowest} possible overlap between reference
and prediction (worst case scenario in fig.~\ref{fig:explops} and lower bound in
fig.~\ref{fig:operator-behaviour}).
\begin{equation}
  \Zf\Luk (\klref_i, \klpred_j) = \max (\klref_i + \klpred_j - 1, 0)
\end{equation}
As $\max (\klref_i + \klpred_j - 1, 0) = \klref_j - min (\klref_i, 1 - \klpred_j)$, $\Zf\Gdl$ and
$\Zf\Luk$ are point symmetric about $(\klpred = \frac{1}{2}; \Zf = \frac{1}{2} \klref)$ to each other
(fig.~\ref{fig:operator-behaviour}).

The matrix diagonal of $\Z\Gdl$ reports the best possible performance that is in accordance
with the given reference and prediction. Likewise, the off-diagonal elements are the worst possible
performance for the respective type of misclassification $\Klref \mapsto \Klpred$. $\Z\Luk$ behaves
antithetically.

Both $\Z\Gdl$ and $\Z\Luk$ lack two properties of crisp confusion matrices that have been identified
as desirable for soft confusion matrices \cite{Pontius2006, SilvanCardenas2008}: firstly, their
marginal sums do not equal the reference and prediction class membership vectors. Secondly, perfect
reproduction does not produce a diagonal confusion matrix and is thus more difficult to recognize
than in crisp classification problems.

Several proposals on how to ``repair'' both marginal sums and diagonal structure of $\Z\Gdl$ for
perfect reproduction of the reference exist \cite{Pontius2006, Pontius2006a,
  SilvanCardenas2008}. They all distribute the remainder of the prediction after the agreement
(diagonal) has been subtracted.  Again, diagonal and off-diagonal elements of these composite
confusion matrices do not share the same interpretation. Silván-Cardenás and Wang
\cite{SilvanCardenas2008} report that the use of soft confusion matrices in practice appears to be
restricted to the diagonal of $\Z\Gdl$, $\Zf\Gdl (\klref_\Kl, \klpred_\Kl)$. 

Interestingly, the properties of $\Z\Gdl$ and $\Z\Luk$ have not yet been interpreted with respect to
resulting bias of the performance measure. Using exclusively the diagonal of $\Z\Gdl$ results in a
strong optimistic bias: only the ``optimistic'' part of $\Z\Gdl$ is used. We are not aware of any
application where the corresponding pessimistic performance measures are reported, although $\Zf\Luk$
as a measure of the least possible overlap is mentioned (but not used) by Pontius \emph{et.al.}
\cite{Pontius2006a}. 

Using a performance measure that has by construction a strong optimistic bias, \ie overestimates the
classifier's performance, is clearly not appropriate for an application where the classifier should
ultimately indicate whether brain tissue is cut out or not. Similar caution is necessary in most
biomedical and many chemometric applications.

We therefore propose to recombine $\Z\Gdl$ and $\Z\Luk$ into ``optimistic'' and ``pessimistic''
confusion matrices $\Z\opt$ and $\Z\pes$ as illustrated in fig.~\ref{fig:optpes}.  These two matrices
hold the best and worst possible performance for the observed test results, and can therefore be
interpreted consistently without the need to distinguish diagonal and off-diagonal
elements. Together, the two confusion matrices span the range of possible performances that is in
accordance with the available reference (gold standard diagnosis) and the observed predictions.  The
ambiguity in the reference labels causes this uncertainty about the true
performance: any performance in this range may be the true performance of the classifier. Due to the
ambiguity or uncertainty in the reference labels, the true performance cannot be further narrowed
down.

We define the performance measures (eqs. \ref{eq:sens}\,--\,\ref{eq:NPV}) so that only single
elements from the diagonal of the confusion matrix $\Z$ are needed to obtain these optimistic and
pessimistic bounds.

The interpretation as best and worst possible performance does not take into account that the
validation results are actually performance \emph{estimates}. ``Best'' and ``worst'' here refer to the
uncertainty due to the ambiguity represented by soft class memberships. The performance estimates are
subject to additional uncertainty due to the sampling of the actual test set and possible instability
of the ``surrogate'' models computed during cross validation etc. However, this is outside the scope
of this paper.

The \emph{product} has been used as \AND-operator for continuous-valued logic as well, \eg in
Reichenbach's probability logic \cite{Reichenbach1935}, and has also been discussed for soft
confusion matrices \cite{Lewis2001,Pontius2006, Pontius2006a, SilvanCardenas2008}:
\begin{equation}
\Zf\prd (\klref_i, \klpred_j) = \klref_i \cdot \klpred_j
\end{equation}
Interpreting the class memberships as probabilities, $\Z\prd$ gives the expected amount of
coincidence for independent processes determining the class memberships. In the mixture
interpretation, $\Zf\prd$ follows from the information loss due to low (spatial) resolution: assume
crisp reference and prediction are available at high resolution, but the location information is lost
(the high resolution data is mixed randomly). The expected confusion matrix (normalized by the
respective number of samples) in this situation is just the product-based confusion matrix
$\Z\prd$. From a Bayesian point of view, a uniform prior is used in both interpretations.

The marginal sums of the product-based confusion matrix $\Z\prd$ behave like the marginal sums of the
crisp confusion matrix, for closed world as well as for one-class classifiers: the row sums are
$\sum_n (\klref \cdot \sum \klpred)$ and the column sums are $\sum_n (\klpred \cdot \sum \klref)$.
The sum over all elements is $\sum_n \sum \klpred \cdot \sum \klref$. Specifically, both marginal
sums and the total element sum of $\Z\prd$ equal the number of samples for closed world
classifiers. However, like the other soft confusion matrices (except $\Z\opt$), $\Z\prd$ is not
diagonal if the prediction equals the soft reference. This may be seen as expression of the remaining
uncertainty or ambiguity arising either from the lack of further information about the (unresolved)
distribution of the classes, or from the uncertainty encoded in both reference and prediction.

\subsection{Calculating the Performance Measures for Soft Reference and Prediction.}
\label{sec:calc}
Eqs.~\eqref{eq:sens} to \eqref{eq:PPV} can directly be used with the soft confusion matrices. Note
that the performance measures refer only to diagonal elements of the confusion matrix $\Z$.  Thus,
all problems due to the marginal sums not equaling prediction and reference membership vectors are
avoided, including possible specificities or negative predictive values \textgreater 100\,\% for
$\Zf\Gdl$. $\Zf\Gdl$ and $\Zf\Luk$ directly yield the most optimistic and most pessimistic
case. Fig.~\ref{fig:suppl-operator} illustrates the performance measures for the three different
operators. Note that each pair of $\Zf\Gdl$ (optimistic) and $\Zf\Luk$ (pessimistic) leaves a quarter
of the input space completely without information: if reference and prediction are too ambiguous,
they are in accordance with any possible value of the performance measure (interval width in
fig.~\ref{fig:suppl-operator}).

For the product operator, the four characteristic measures simplify to prediction, 1 - prediction,
reference and 1 - reference for each sample and weighted averages thereof for the multi-sample
performance.

\paragraph{The Difference between Prediction and Reference} 
The mixture interpretation of soft memberships suggests a treatment of soft
classification analogous to regression.

Regression residuals measure the deviation of the prediction from the reference. A short inspection
of the soft confusion matrices reveals that $\Zf\Gdl$ yields performance 1 if the prediction
$\klpred$ equals (or exceeds) the reference $\klref$. This means that it inherently reports
deviations from the ground truth or gold standard diagnosis (though only for too low estimates, too
high estimates are not penalized).  $\Zf\Luk$ produces a more complicated behavior (bottom row of
fig.~\ref{fig:suppl-operator}).

In contrast, the resulting performance measures for the product operator simplify to the regression
errors distributed according to the reference memberships: analogous to the calculation of regression
residuals $\varepsilon = \hat y - y$, we compare the observed confusion matrix $\Z\prd (\klref,
\klpred)$ with the ``ideal'' confusion matrix for the actual reference $\Z\prd (\klref, \klpred =
\klref)$ \cite{Lewis2001}:
\begin{equation}
  \label{eq:delta}
  \D\prd = \Z\prd (\klref, \klpred) - \Z\prd (\klref, \klref)
\end{equation}
Just as for regression, the sign of the residuals distinguishes over- or underestimation. Thus we sum
the absolute deviations rather than their signed values (squared deviations are discussed below). In
closed world systems, every underestimation implies the same amount of overestimation in other
classes: the row sums of $\D\prd$ are 0.

Also, $\D$ measures an error, so we compute the complementary $1 - |\D|$ as our performance measures
refer to the correct part of the prediction:
\begin{align}
   \Sens\mae (\klref, \klpred) &= 1 - \frac{\sum_n |\D\prd (\klref, \klpred)|}{\sum_n \klref} \\
  &= 1 - \frac{\sum_n |\Zf\prd (\klref, \klpred) - \Zf\prd (\klref, \klref)|}{\sum_n \klref} \\
  \label{eq:sensdelta} & =1 - \sum_n \frac{\klref}{\sum_n \klref}~ |\klpred - \klref| 
\end{align}
$\Sens\mae$ uses the mean absolute error weighted by the reference memberships (compare
eq.~\ref{eq:sens}).  The specificity reports the remainder of the residuals, which is attributed to
samples not belonging to the class:
\begin{align}
  \Spez\mae (\klref, \klpred) &= 1 - \sum_n \frac{1 - \klref}{\sum_n 1 - \klref}~ |\klpred - \klref|
\end{align}

The predictive values characterize the inverse thought, consequently deviations are distributed
according to the \emph{predicted} memberships:
\begin{align}
  \PPV\mae (\klref, \klpred) &= 1 - \sum_n \frac{\klpred}{\sum_n \klpred}~ |\klpred - \klref|\\
  \NPV\mae (\klref, \klpred) &= 1 - \sum_n \frac{1 - \klpred}{\sum_n 1 - \klpred}~ |\klpred - \klref|
\end{align}

\paragraph{Mean Absolute Error MAE and Root Mean Squared Error RMSE} Instead of the weighted MAEs, the
respective RMSEs can be used, \eg:

\begin{align}
  \Sens\rmse (\klref, \klpred)
  & =1 - \sqrt {\sum_n \frac{\klref}{\sum_n \klref}~ (\klpred - \klref)^2} 
\end{align}
The MAE is more closely related to the usual error counting for crisp classifiers, while the RMSE is
more common for regression models. For calculating the performance of soft prediction and crisp
reference, the mean squared error MSE is also known as Brier score \cite{Brier1950}.

\begin{figure}[tb]
  \centering \includegraphics[width=0.75\hlw]{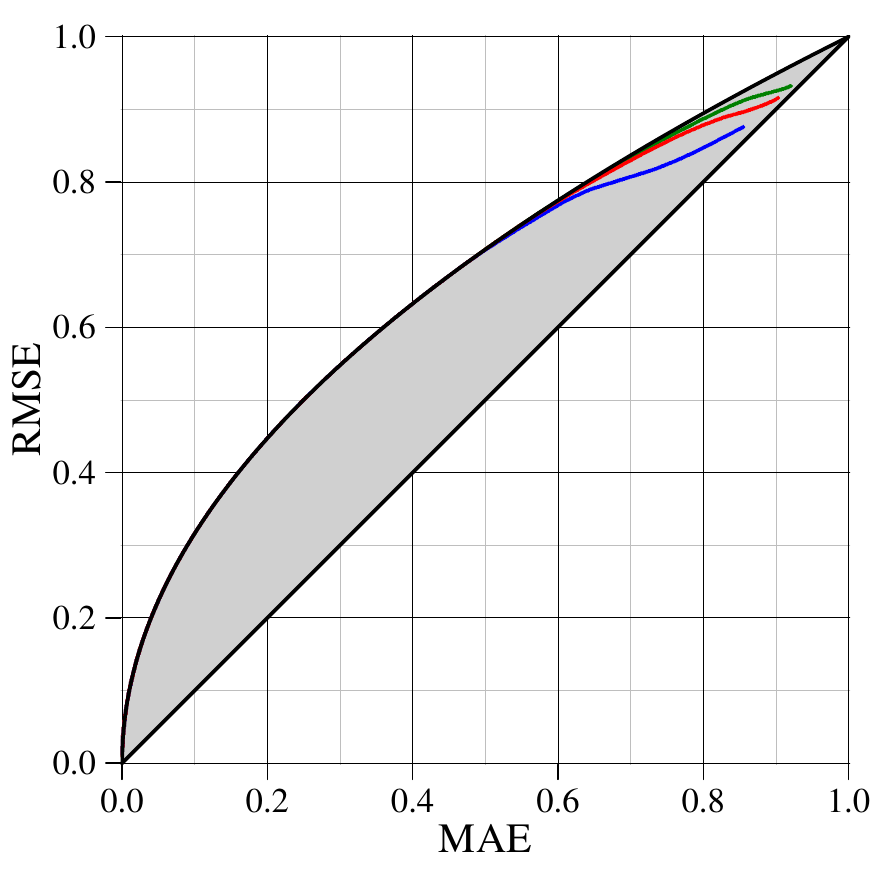}
  \caption{Bounds of the RMSE for crisp reference data (black, shaded). Above an MAE of 0.6, the soft
    references further restrict the range MAE and RMSE can possibly take and upper bounds for the
    sensitivity of the tumor classes N (green), A~\grade II (blue), and A~\grade III+ (red) deviate
    from the black maximum RMSE for crisply labelled data.}
  \label{fig:rmsebound}
\end{figure}

$\MAE$ and $\RMSE$ are related: In general, $\MAE \leq \RMSE \leq \sqrt n \MAE$ with the respective
number of samples $n$. For classification, however, no single prediction can deviate by more than $1$
from the reference (and this only for crisp reference memberships): $0 \leq \MAE \leq 1$. Thus, $\MAE
\leq \RMSE \leq \sqrt{\MAE}$. For soft reference, the upper bounds of both MAE and RMSE are lower, as
the maximal deviation is the greater of $\klref$ and $1 - \klref$, respectively, for each
sample. Fig.~\ref{fig:rmsebound} illustrates the bounds for crisp reference data as well as for our
application.

\paragraph{Inter-class performance}
$\D\prd$ and the derived performance measures count both under- and overestimations.  This yields the
expected behavior for class-wise performance measures. Performance measures that summarize more than
one class (\eg overall accuracy) should either take care of the consequences beforehand, or they may
be normalized according to the maximal possible error. Closed world classifiers have one under- and
one overestimation for each misclassification, thus $\MAE \leq 2$ and $\RMSE \leq \sqrt{2}$. For
one-class classification the bounds are $\MAE \leq \nkl$ and $\RMSE \leq \sqrt{\nkl}$.

\subsection{Implementation and Availability}
We implemented the proposed performance measures in R \cite{R} as package ``softclassval''. The
package is released under GPL 3 (\url{http://www.gnu.org/licenses/gpl.html}).

The project is hosted at \url{http://softclassval.r-forge.r-project.org} where the current
development version, its check results and the source of previous versions (via the version control
web interface) are available. The checks include unit tests to ensure calculational correctness,
which consist of ca. twice as many lines of code than the actual function
definitions. \texttt{softclassval.unittest ()} executes the unit tests in interactive R sessions if
package svUnit \cite{svUnit} is available.

Stable releases can conveniently be installed from the Comprehensive R Archive Network CRAN
(\url{http://cran.r-project.org/package=softclassval}, both binaries and source code are available) by
executing \texttt{install.packages ("softclassval")}. Check results from CRAN for a variety of
platforms can be inspected at
\url{http://cran.r-project.org/web/checks/check_results_softclassval.html}.

\section{Application to Astrocytoma Grading}
\label{sec:appl-astr-grad}

\subsection{Experimental and Data Analysis Set-Up}
\label{sec:exper-data-analys}
\begin{figure*}[tb]
\subfloat[\label{fig:sample:hf}bulk sample]{\includegraphics[height=3.25cm]{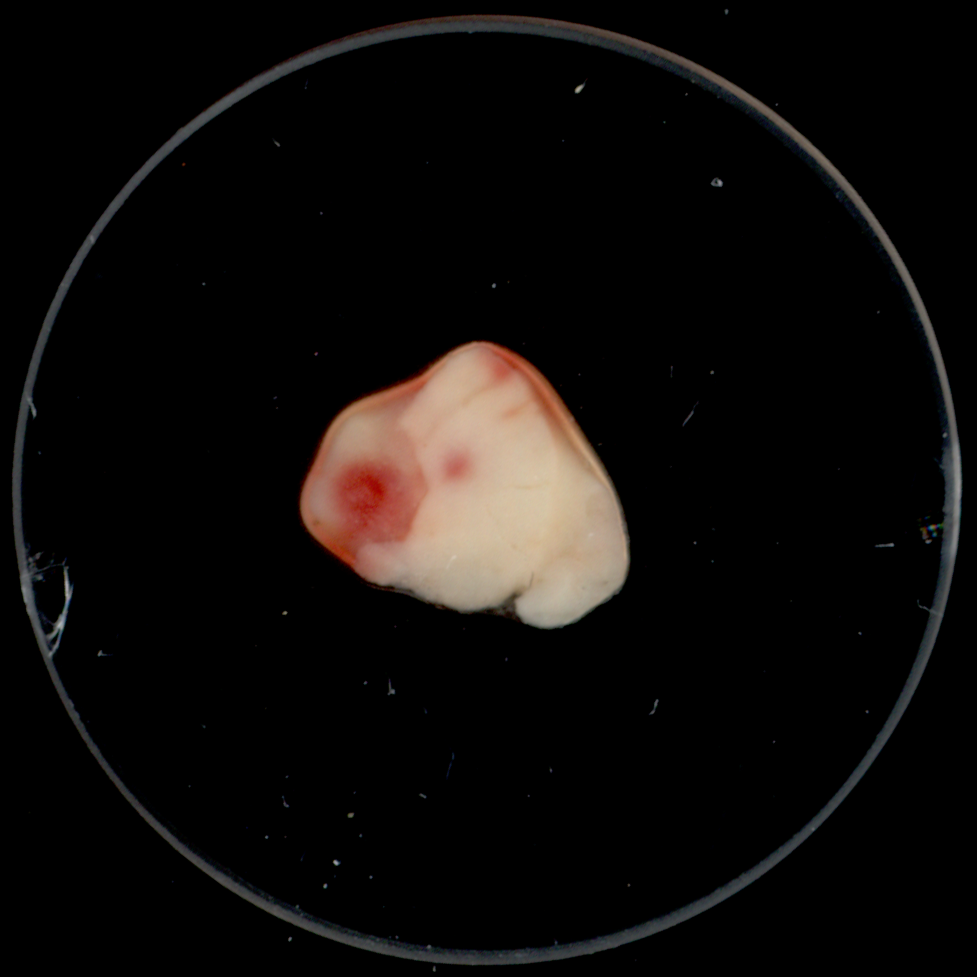}}\hfill%
\subfloat[\label{fig:sample:diag}detailed diagnosis]{\includegraphics[height=3.25cm]{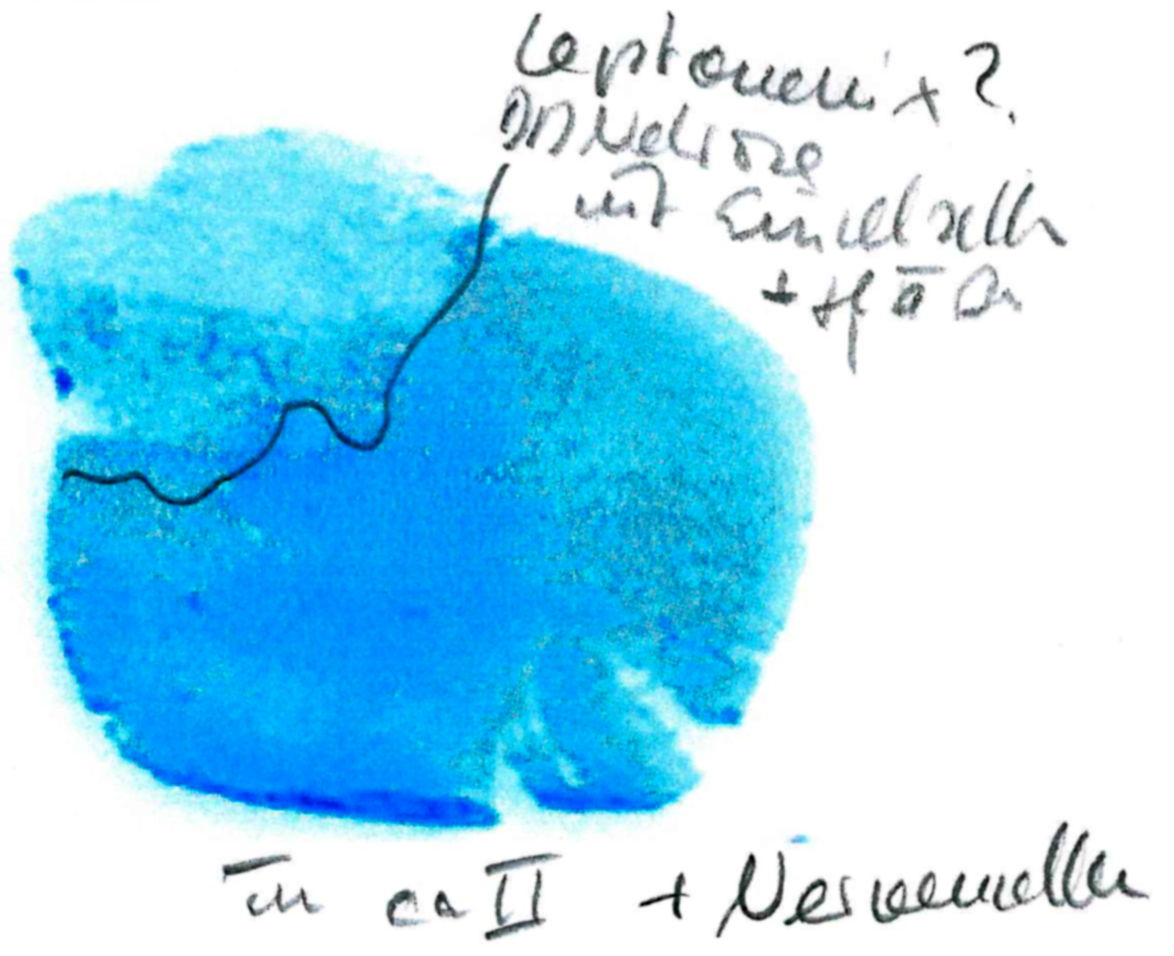}}\hfill%
\subfloat[\label{fig:sample:ref}reference labels]{\includegraphics[height=3.25cm]{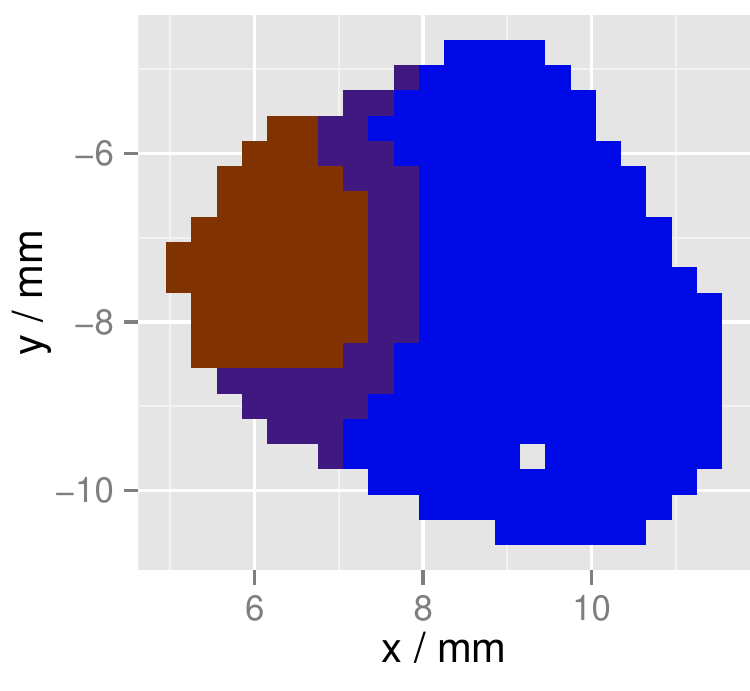}}\hfill%
\subfloat[\label{fig:sample:pred}predictions]{\includegraphics[height=3.25cm]{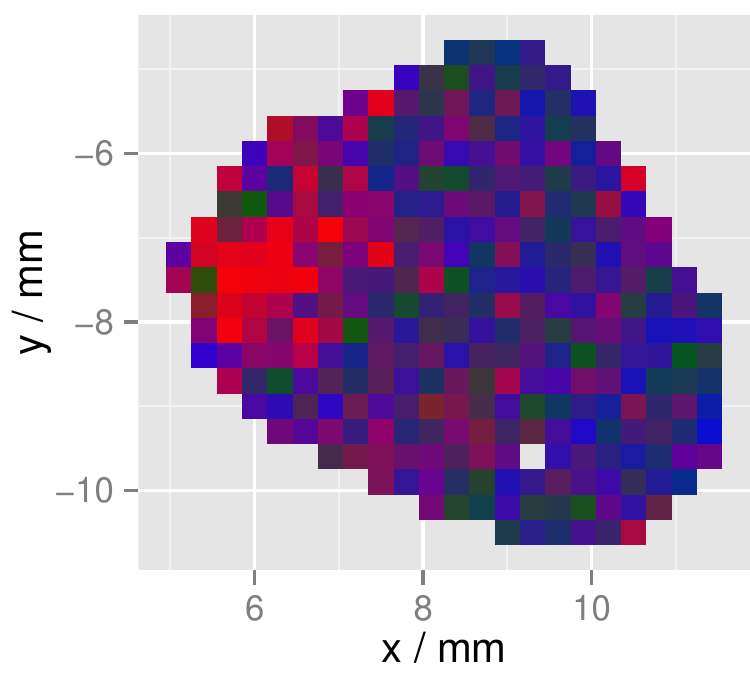}}\hfill%
\subfloat[\label{fig:sample:legend}legend]{\includegraphics[height=3.25cm]{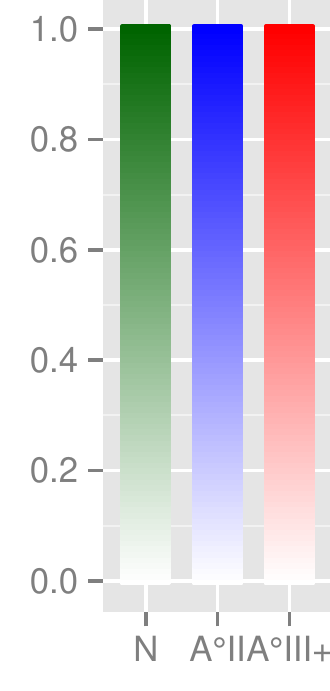}}
\caption{Samples. From left to right: \protect\subref{fig:sample:hf} bulk sample ready for measurement;
  \protect\subref{fig:sample:diag} histology results for methylene-blue stained parallel section;
  \protect\subref{fig:sample:ref} reference labels; \protect\subref{fig:sample:pred} predictions from one of the 125
  cross validation iterations. The colors for reference labels and prediction are obtained by mixing
  the respective parts of the green (class N), blue (class A\grade II) and red (class A\grade III+)
  colours which stand for the pure tissues as indicated in the legend
  \protect\subref{fig:sample:legend}. Most of the sample area is diagnosed as a \emph{mixture} of tumor
  \grade II with normal cells (reference membership: 0.1 N, 0.9 A\grade II, 0.0 A\grade III+; blue
  color). On the left is an area where the pathologist expressed \emph{uncertainty}: the tissue may
  be either normal leptomeninges or necrotic. This uncertainty was translated to reference labels of
  0.5 N, 0.0 A\grade II and 0.5 A\grade III+ (brown). A transition zone between these areas got
  intermediate reference labels (dark violetish). The cross validation results show some noise and
  decidedly lean towards necrosis for the area where the neuropathologist was uncertain.}
  \label{fig:sample}
\end{figure*}

\paragraph{Experiments and Reference Labels}
We prepared cryo sections of our samples which were stained for reference diagnosis (for the classes,
see classifier setup below). Raman maps were recorded of the adjacent side of the remaining bulk
tissue on an evenly spaced grid with step sizes between 200 and 333\mum using a fiber-optic probe
with focus diameter of ca. 60\mum (order of magnitude: 10\textsuperscript{3} cells). Figure
\ref{fig:sample:hf} shows such a bulk sample immediately before Raman measurements.

Histological diagnosis was obtained for the parallel section (fig.~\ref{fig:sample:diag}) and
transferred to the measurement grid \emph{without} any display of the spectra
(fig.~\ref{fig:sample:ref}).  Partial class memberships were used for the reference labels where
ambiguity or uncertainty occurred. For example, tumor tissue between the classes (``A\grade II to
\grade III'') was labeled belonging half and half to the respective classes. The diagnosis
``individual tumor cells in normal tissue'' and tissue where the histologist was not sure whether it
contained tumor cells were labeled as 0.05 tumor and 0.95 normal, and so forth.  If shape or deformation
of the sample prevented the transfer of the diagnosis, the fractions of the respective areas on the
reference section were used as class membership.

\begin{table}[tb]
 \caption[]{Overview of the data set \cite{Beleites2011}. With kind permission of Springer Science+Business Media.}
 \label{tab:daten}
 \centering
 \begin{small}
\begin{tabular}{lrrrr}
  \hline
  & \multicolumn{2}{c}{crisp reference}  & \multicolumn{2}{c}{crisp + soft reference} \\
class & patients & spectra  & patients & spectra \\\hline
Normal & 16 & 7\,456 & 35 & 15\,747 \\
 ~~thereof controls & 9 & 4\,902 & 9 & 4\,902 \\
 Astrocytoma \grade II & 17 & 4\,171 & 47 & 19\,128 \\
 \vspace{0.5ex}Astrocytoma \grade III+ & 27 & 8\,279 & 53 & 21\,617 \\
 total & 53 & 19\,906 & 80 & 37\,015 \\\hline
\end{tabular}
 \end{small}
\end{table}
\begin{figure}[tb]
 \centering
 \includegraphics[width=\linewidth]{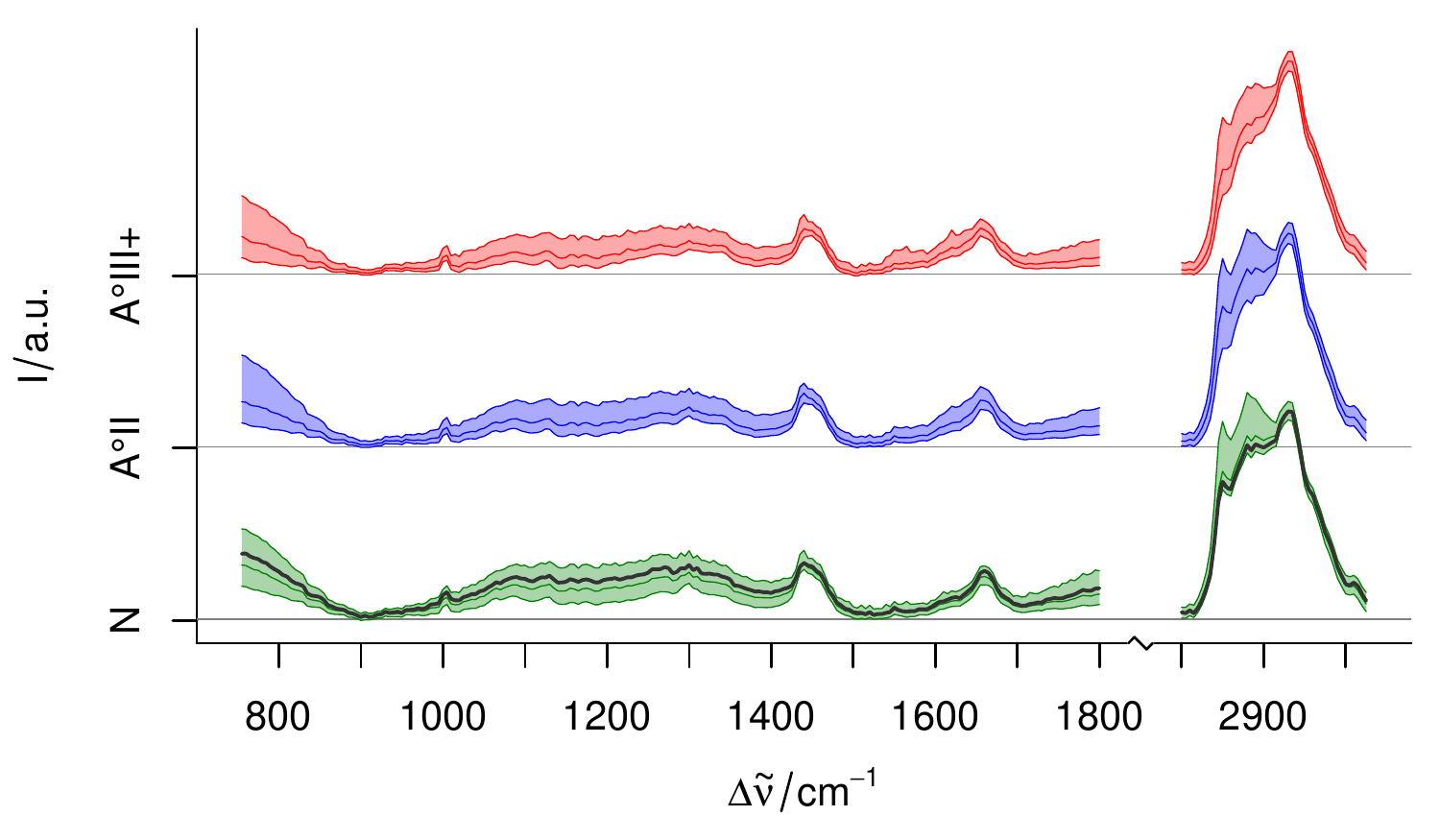} 
 \caption[]{Weighted median, 16\th, and 84\th percentile spectra. Thick line: the mean grey tissue
  spectrum used for centring \cite{Beleites2011}. With kind permission of Springer Science+Business Media.
}
 \label{fig:spectra}
\end{figure}
Figure~\ref{fig:spectra} gives an overview of the pre-processed spectra immediately before centering
on the average spectrum of normal grey matter. Table~\ref{tab:daten} summarizes the data set. A more
detailed discussion of the samples and data set as well as spectroscopic interpretation have been
given in \cite{Beleites2011}.

\paragraph{Classifier Set-Up}
The samples include a number of different tissues that were combined into the three classes requested
by the neurosurgeons:
\begin{description}
\item[N (normal or non-tumor tissues):] normal white matter, normal gray matter, and small amounts
  of gliotic tissue. Surgically, such tissue \emph{must} be preserved. For convenience, we refer to
  this class as ``normal'' in the text.
\item[A\grade II (low grade tumor morphology):] such tissue would lead to a diagnosis of an
  astrocytoma \grade II if it is the most de-differentiated tissue found. Surgically, this may be
  thought of as ``take out if possible''.
\item[A\grade III+ (high grade morphology):] malignant or high grade tumor tissues, comprising
  \grade III and IV morphologies as well as necrotic tissue. These tissues \emph{must} be excised.
\end{description}
Note that both class boundaries are of practical importance: the boundary between normal and
low grade tissues is the intended excision border. Yet, in order not to risk damage to normal
brain tissue, the surgeons frequently have to back up to the border between low and high grade
morphology.

The three classes are ordered with increasing malignancy. Nevertheless, we model unordered classes
here, and an extension to ordered class models is outside the scope of this paper.
We use here the same soft LR classifier as in \cite{Beleites2011}. Briefly, the classifier was
trained using the R \cite{R} package nnet \cite{Venables2002}. All pre-processing was decided by
spectroscopic knowledge, no data-driven steps were included and no parameter optimization was
performed.  However, we checked that a PLS projection \cite{pls} of the spectra onto 25 latent
variables as pre-processing for LR training did not lead to more than slight changes in the
prediction (see supplementary figure~\ref{fig:pls-preproc}).  90\,\% of the PLS-preprocessed
predictions lie within $\pm$ 0.07 of the respective predictions without PLS pre-processing. As a
comparison, 90\,\% of the differences between different cross validation iterations for the same
spectrum are within $\pm$ 0.2. The root mean squared difference between iterations is 2.7$\times$ the
root mean squared difference for PLS-pre-processing with 25 latent variables.

\paragraph{Validation}
125 iterations of an 8-fold cross validation scheme were used, splitting the data patient-wise since
spectra of one patient are not statistically independent.

Our data set does not reflect the relevant prior probabilities, nor are any such data
available. Therefore, we calculate only sensitivity and specificity and do not report predictive
values.

\paragraph{Software} Data analysis was performed in R \cite{R} using the packages R.matlab
\cite{R.matlab} for data import, hyperSpec \cite{hyperSpec} for spectra handling, pls
\cite{Mevik2007} for multiplicative signal correction of co-additions of the spectra and the PLS
pre-processing for comparison, nnet \cite{Venables2002} for the logistic regression, and ggplot2
\cite{ggplot2} for graphical display.

\subsection{Results of the Astrocytoma Grading}
\label{sec:results-discussion}
We report corresponding triples of one performance measure of all three classes separated by bars
(N\,|\,A~\grade II\,|\,A~\grade III+). A tabular overview of the results is available in the
supplementary material \textbf{tab.~\ref{tab-results}}.

\paragraph{Best, Expected, and Worst Case Performance}

\begin{figure*}[tb]
  \subfloat[\label{fig:prodops-lr:observed}Soft \AND-operators]{\includegraphics[width=.33\linewidth]{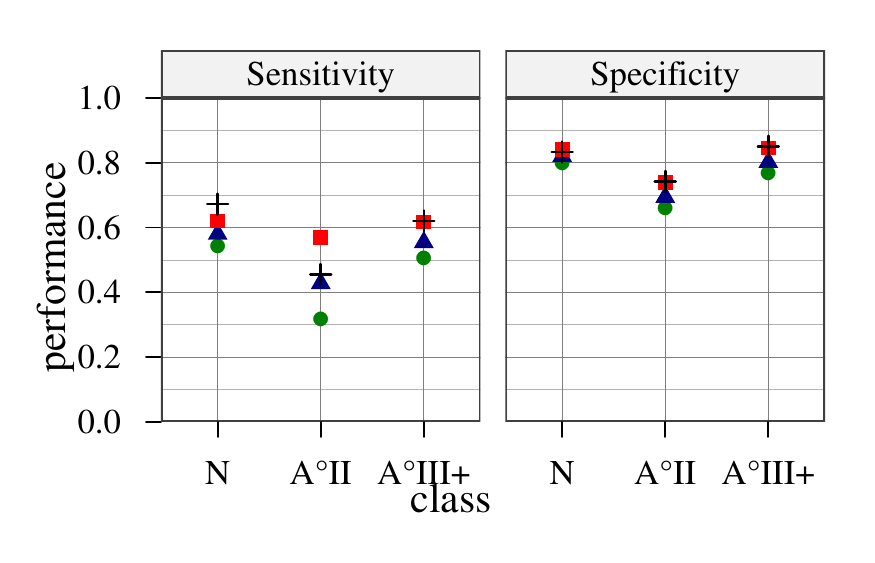}}
  \subfloat[\label{fig:prodops-lr:ideal}Ideal Performance]{\includegraphics[width=.33\linewidth]{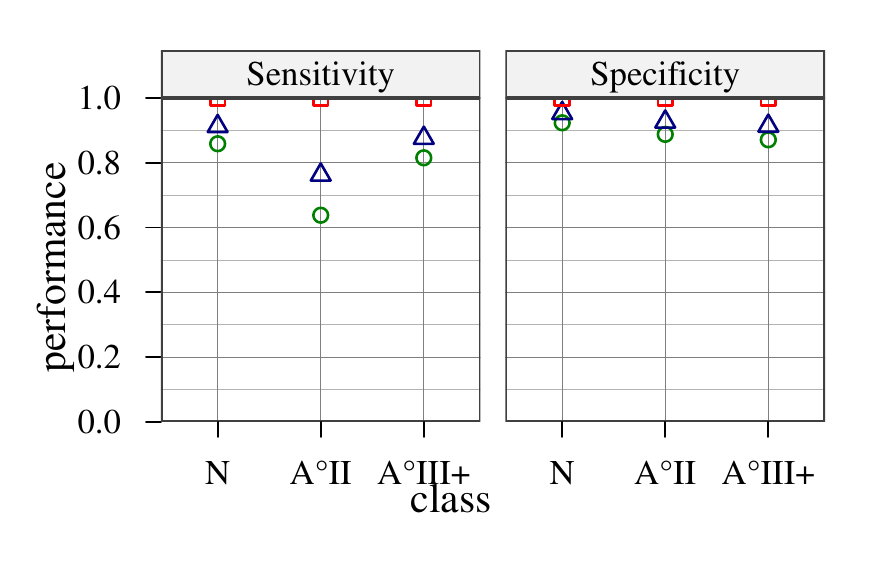}}
  \subfloat[\label{fig:deltaops-lr}Regression-like performance measures]{\includegraphics[width=.33\linewidth]{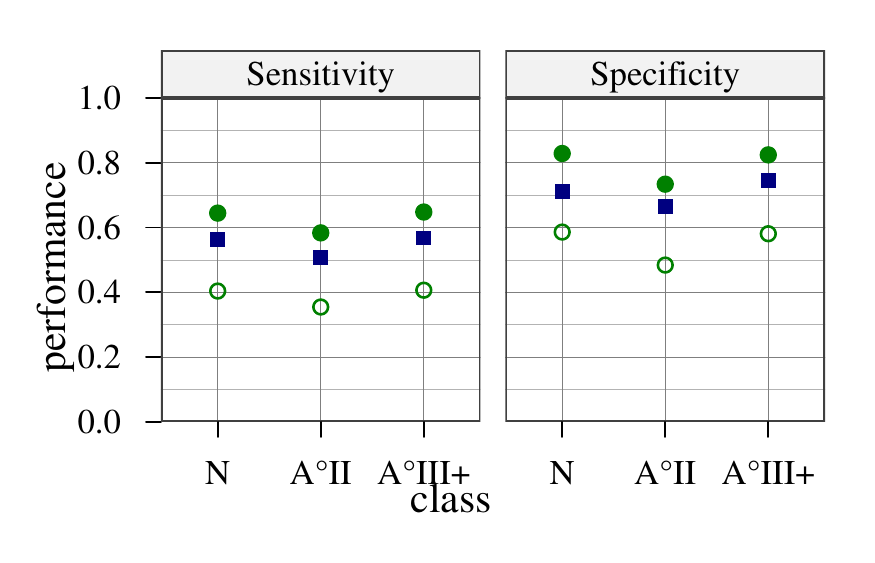}}
  \caption[]{Results for the astrocytoma grading. \subref{fig:prodops-lr:observed} Expected (product;
    blue triangle), best (weak \AND; red square) and worst case (strong \AND; green circle) overlap
    between predicted and reference averaged over all spectra and iterations. Black crosses mark the
    (coinciding) results of all three operators for the crisp reference
    spectra. \subref{fig:prodops-lr:ideal} The (hypothetical) results for ideal reproduction of the
    reference labels is shown (open symbols).  \subref{fig:deltaops-lr} Performance based on the
    difference between prediction and reference: $1 - \wMAE$ (circles) and $1 - \wRMSE$
    (squares). The open circles are $1 - \sqrt{\wMAE}$, the lower bound of the $\wRMSE$ given the
    observed $\wMAE$.}
  \label{fig:prodops-lr}
\end{figure*}

Figure~\ref{fig:prodops-lr:observed} shows the expected (product \AND), best (weak \AND) and worst
case (strong \AND) sensitivity and specificity of our models. Note that this range accounts solely
for the ambiguity of the reference data. It does not account for the the uncertainty due to the
number of test cases nor for uncertainty due to model instability.  While these uncertainties are not
a topic of the present study, it may be noted that the standard deviations of the performance
measures observed over the 125 iterations of the cross validation range from 0.007 to 0.013 for the
sensitivities and from 0.004 to 0.006 for the specificities. For the unambiguously labeled (crisp)
samples, standard deviations between 0.005 and 0.017 were observed. All these are much smaller than
the symbol sizes in fig.~\ref{fig:prodops-lr:observed}.

The expected sensitivity for the intermediate tissue morphology A~\grade II, 0.43, is lower than the
sensitivity for both normal (0.58) and high grade (0.55) morphologies. This corresponds to the
A~\grade II class also biologically being in between normal and high grade, \ie the class has two
borders relevant to the classification problem whereas normal and high grade classes have only one
relevant border. This pattern is even stronger for the strong sensitivity (0.54\,|\,0.32\,|\,0.50), but
almost vanishes for the weak sensitivity (0.62\,|\,0.57\,|\,0.62).

A similar overall pattern is observed for the expected specificities (0.82\,|\,0.69\,|\,0.80). Normal and
high grade tissue are rarely confused, the difficulties in the classification lie between the
consecutive classes.

The difference between strong and weak \AND largely reflects the amount of ambiguity in the reference
labels. This becomes clear by comparison with fig.~\ref{fig:prodops-lr:ideal}, where the overlap for
ideal reconstruction of the reference data is shown.  The more ambiguous the reference, the larger
the gap between weak and strong performance measure: given the reference labels, the expected
sensitivity for A\grade II cannot exceed 0.76, whereas for N and A\grade III+ 0.91 and 0.88 could be
reached.  The strong sensitivities (lower bound) cannot be more than 0.86\,|\,0.64\,|\,0.82 for the three
classes. The specificity is calculated with all samples that do \emph{not} belong to the class in the
denominator, and has therefore less ambiguity. Thus expected specificities of up to 0.95\,|\,0.93\,|\,0.91
and worst-case specificities of up to 0.92\,|\,0.89\,|\,0.87 for the three classes are possible. The weak
sensitivity and specificity can always reach 1.

The A\grade II class has soft borders to both other classes, whereas there is much less ambiguity in
the reference labels between normal and high grade tissues. Class N \emph{references} are less
ambiguous than the high grade morphologies A\grade III+ (fig.~\ref{fig:prodops-lr:ideal}). In
contrast, the \emph{predictions} with respect to class N reach about the same sensitivity and
specificity as those of class A\grade III+ (fig.~\ref{fig:prodops-lr:observed}).

\begin{figure}[tb]
  \includegraphics[width=0.67\hlw]{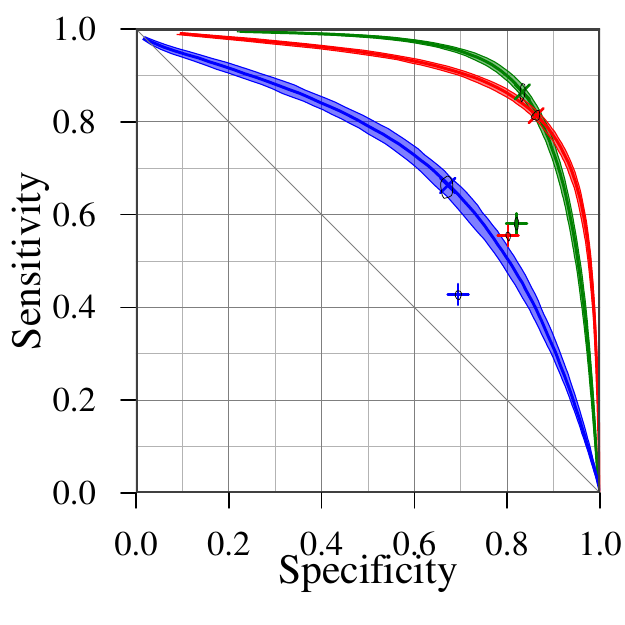}
  \caption{Specificity-sensitivity-diagram for hardened predictions of the crisp data. $\times$:
    hardening with threshold $\frac{1}{3}$.  $+$: soft performance. As only crisply labelled spectra
    are included, all soft \AND-operators coincide.  Colors: green: N, blue: A\grade II, red: A\grade
    III+. The bands show the inter quartile range over the 125 cross validation iterations, thick
    line: median; The ROC-like lines are produced by varying the hardening threshold. Hardening
    discards small deviations that are penalized by the soft measures: $+$ are below the
    specificity-sensitivity-curves.\\
    The contours behind each $\times$ and $+$ contain ca. 50\,\% of the observed values over the 125
    iterations: hardening considerably increases the variance.}
  \label{fig:prodops-roc}
\end{figure}

Fig.~\ref{fig:prodops-roc} compares the behaviour of the commonly used (crisp) sensitivity and
specificity with the soft \AND-operators for the crisp spectra.  The specificity-sensitivity curves
of the three classes were calculated using the R \cite{R} package ROCR \cite{ROCR}. The band width
illustrates the variation in model performance due to different composition of the training data in
different iterations of the cross validation: shown are the inter quartile range and median (25\th,
50\th and 75\th percentiles).

The primary output of the logistic regression models are posterior probabilities. As a post-processing
step, the spectrum can be assigned to a class if the respective posterior probability exceeds a given
threshold (hardening the prediction). To obtain the specificity-sensitivity lines, this threshold is
varied. 

Hardening discards small deviations that do not cross the threshold, which are therefore not detected
in the specificity-sensitivity-curve. In contrast, the soft \AND-operators work directly on the
posterior probabilities and penalize already small deviations from the reference. This leads to the
soft performance values (marked $+$) lying below the specificity-sensitivity-curves. The classical
calculation of specificity-sensitivity-curves is possible only for crisply labelled spectra. In order
to have the same test sample basis throughout the graph, soft spectra were therefore excluded from
all calculations for figure~\ref{fig:prodops-roc}, \ie the soft performance measures are the same
results marked by plus (+) signs in figure~\ref{fig:prodops-lr:observed}. For crisp reference, all
three \AND-operators yield the same result as there is no ambiguity (see also
fig.~\ref{fig:operator-behaviour}).

Hardening also influences the variance of the performance measure.  On the one hand, if the hardening
threshold is in the range of the predicted posterior probabilities, hardening will increase the
variance on the performance measure. This causes the well-known high variance of the crisp classifier
performance measures: testing with crisp class labels is described as a Bernoulli-process, leading to
the variance of the observed performance$\sigma^2 (\hat p) = \frac{p (1 - p)}{n}$. The soft
performance measures do not suffer from this increase in variance. The increase in variance will
usually be high for models that have rather gradual transitions between the classes.  On the other
hand, for such a model extreme thresholds will mean that (almost) all samples are on the same side of
the threshold. In this (far less common) case, hardening actually lowers the variance. In our
example, that would be e.g. if the A\grade II class were operated at sensitivity of 0.95 with a
specificity of 0.10. Operating a classifier at such an extreme working point actually requires an
adapted training strategy, including also an adequate composition of the training set.

Note, however, that extreme thresholds for models that in fact predict intermediate posterior
probabilities are fundamentally different from models with very sharp class transition: if the
transition between the classes is immediate, hardening has hardly any effect on the variance, as the
predicted posterior probabilities are already close to 0 or 1.

In our data, assigning each test sample completely to the class with the highest posterior
probability (threshold $\frac{1}{\nkl} = \frac{1}{3}$, working points marked $\times$), we observe
crisp sensitivities and specificities of (0.86\,|\,0.66\,|\,0.81) and (0.83\,|\,0.67\,|\,0.86),
respectively -- a typical choice of working points close to the major diagonal of the
specificity-sensitivity-diagram.  The corresponding variances of the soft sensitivities and
specificities (marked $+$) are (1.64\,|\,0.64\,|\,0.52)~ $\times 10^{-4}$ and
(0.16\,|\,0.36\,|\,0.21)~$\times 10^{-4}$ whereas the variances of the crisp performance measures are
(2.84\,|\,4.13\,|\,0.85)~$\times 10^{-4}$ and (0.31\,|\,2.17\,|\,0.68)~$\times 10^{-4}$. In other
words, this default hardening  increases the variance about 60 -- 550\,\%  (sensitivity of A\grade
III+ and A\grade II, respectively).

Hardening can thus be seen as a noise reduction technique that can be beneficial for the predictive
performance. However, for the measurement of performance, information is lost.  This loss becomes
important in optimization of classifiers: the optimizer will not be able to distinguish well between
slightly different models using any target function on the basis of hardened predictions. Even worse,
the most difficult class, A\grade II, is most affected by this avoidable increase in variance.

Figure~\ref{fig:deltaops-lr} gives the results for the calibration-type performance measures. These
follow the same general patters already discussed for the direct application of the different
\AND-operators: specificity is higher (0.65\,|\,0.58\,|\,0.65; $1 - \wMAE$) than sensitivity
(0.83\,|\,0.73\,|\,0.82) and the low grade morphologies are the most difficult class. Again, the standard
deviation observed over the 125 iterations is lower than the symbol size: between 0.004 and 0.013 for
$1 - \wMAE$ and between 0.006 and 0.012 for $1 - \wRMSE$.
All $1 - \wRMSE$ but the specificity for normal tissue are closer to the $1 - \wMAE$ than to the $1 -
\sqrt{\wMAE}$. This indicates small deviations from the reference for many samples rather than few
grossly misclassified samples.

\section{Summary and Conclusions}
\label{sec:conclusions}
We propose a set of performance measures (sensitivity, specificity, predictive values, hit- or error
rates, etc.) that can be calculated for classifiers with continuous outcome without the need of
``hardening''. 

Ambiguity of the reference of cases that are borderline according to the ground truth or gold
standard diagnosis leads to ambiguity in the measured performance as well. The proposed measures
reflect this as worst case (strong \AND), best case (weak \AND) and expected (product \AND)
performance.

Deviations from the reference can also be evaluated using weighted versions of well-known calibration
performance measures, namely the weighted mean absolute error $\wMAE$ and the weighted root mean
squared error $\wRMSE$. Their comparison in addition allows to distinguish situations with many
small deviations from few large errors.

Our new  measures improve their classical counterparts in four different ways:

\begin{enumerate}
\item While classification assumes perfectly distinct classes, in reality this is often not the
  case. In the past, samples with ambiguous reference labels (borderline cases) usually were
  excluded completely from both classifier training and testing, or their reference labels were
  hardened. This has serious consequences. Excluding borderline cases from classifier training can
  lead to overestimation of class separability, while hardening of class labels samples truly in between
  the classes (\eg mixed cell population or cell population currently undergoing de-differentiation) will
  actually drive the model to overestimate class separability.  However, as the classical performance
  measures do not allow to evaluate the model performance for borderline cases, this overoptimistic
  modeling of class separation could not be detected, the same is true if the reference labels were
  hardened. If the predictions are hardened as well, neither can under-estimation of class
  separability be detected. Excluding borderline cases leaves the validation completely blind for the
  behaviour of the classifier close to the class boundaries defined by the reference. Hardened
  reference labels probe this region, but high variance results. In contrast, the soft
  performance measures penalize over- or underestimation of class separability and have lower variance than
  their hardened counterparts. They thus open the way for more realistic modeling of class boundaries
  that uses also borderline training cases.

\item Truly ambiguous samples may be the actual target of a classifier, such as in our example of
  astrocytoma grading for surgical guidance.  In that case, borderline cases are \emph{most
    important} test samples, since testing clear cases only cannot be considered representative for
  the application.  The proposed soft measures work with samples diagnosed as borderline cases and
  thus allow for more realistic classifier testing.

\item Hardening of the outcome has (like other dichotomization approaches) been criticized due to the
  inherent loss of information. This causes difficulties when classifier performance is
  compared. Model optimization relies on detecting already small differences in the predictive
  ability of the models which is thwarted by the hardening. In contrast, the soft performance
  measures already report small deviations from the reference and thus allow to differentiate between
  more similar models than their crisp counterparts.

\item For models with gradual class transitions (i.e. that actually predict intermediate posterior
  probability values), typical hardening thresholds lead to an increase in variance that is avoided by the
  soft performance measures. For our astrocytoma grading, the sensitivity and specificity based on
  the soft \AND-operators show between 39 and 84\,\% less variance over the 125 iterations of the
  8-fold cross validation than sensitivity and specificity based on crisp classification.
\end{enumerate}

\section*{Acknowledgments}
Financial support by the Associazione per i Bambini Chirurgici del Burlo (IRCCS Burlo Garofolo
Trieste) is highly acknowledged.

\bibliographystyle{elsarticle-num}
\bibliography{Literatur-softerrors}

\processdelayedfloats
\clearpage
\appendix
\setcounter{section}{19}
\setcounter{figure}{0}
\setcounter{table}{0}
\renewcommand{\appendixname}{}
\section*{Supplementary Material}%
\begin{figure*}[h!]
\includegraphics[width=\linewidth]{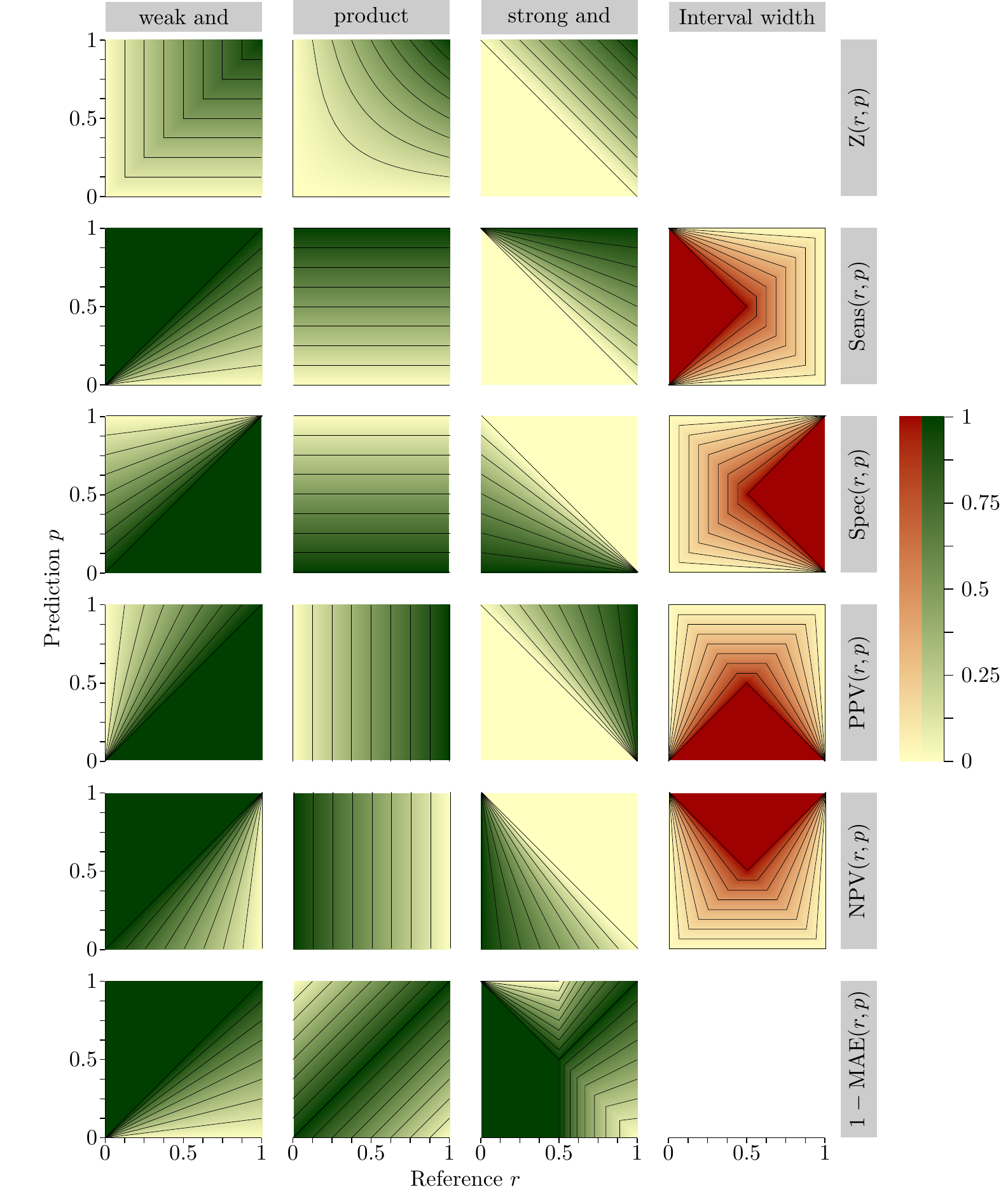}  
  \caption{The values (color) of the three operators (columns) and the soft performance measures
    (rows) for a single sample as function of reference and prediction.  The 4\th column (red) gives
    the width of the interval between strong and weak measures. The interval ranges from 0 to 1 for
    the triangle between the side where the crisp measure is not defined and the center of the input
    space $(\klref = 0.5; \klpred = 0.5)$. Note the symmetry between the performance measures
    (compare also fig.~\ref{fig:symmetrie}). $\Zf\prd$ are similar to $\Zf\Gdl$ for small values and
    similar to $\Zf\Luk$ for high values.\\
    The last row gives the $\MAE$-version of the sensitivity.}
  \label{fig:suppl-operator}
\end{figure*}
\processdelayedfloats
\begin{figure*}[h!]
  \includegraphics[width=.5\linewidth]{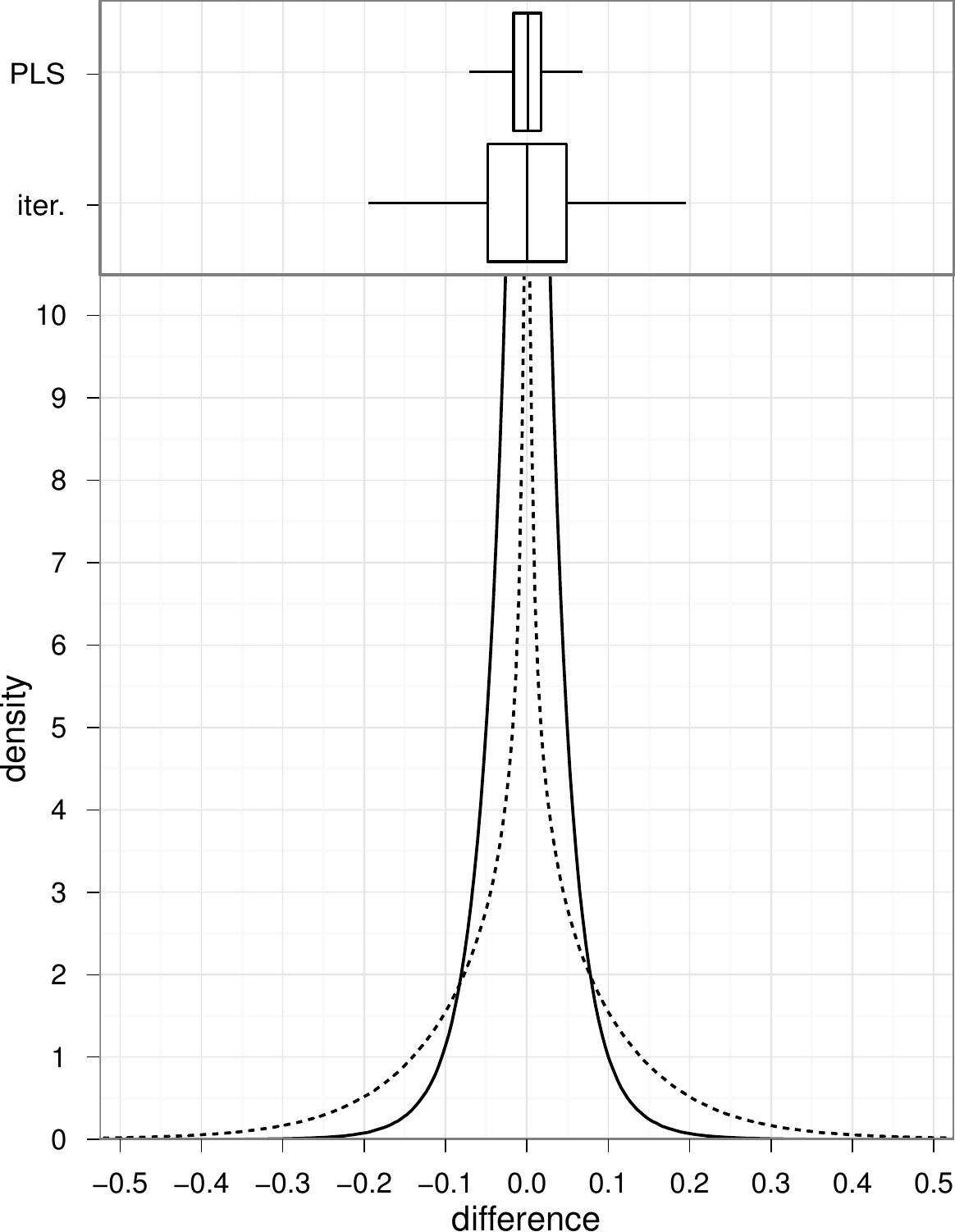}
  \caption{Distribution of differences in the prediction between classifiers with and without 25 LV
    PLS as pre-processing (LR vs. PLS-LR, same iteration). To better compare the results, the
    distribution of differences between the iterations of the LR (without PLS) are shown as well
    (dotted) line.  90\,\% of the PLS-preprocessed predictions lie within $\pm$ 0.07 of the
    respective predictions without PLS pre-processing, whereas the 5\th to 95\th percentile of the
    between-iteration differences range from -0.2 to +0.2.  }
  \label{fig:pls-preproc}
\end{figure*}
\processdelayedfloats
\begin{table}[h]
\caption{\label{tab-results}Results for the Astrocytoma Grading using soft LR.}
\subfloat[\label{tab-results-and} The different \AND-operators. The last three columns (``ideal'') give the best possible  performance that can be obtained with the given reference, i.\,e. the result if the prediction equals the reference memberships.]{
\begin{tabular}{lllSSSSSSS}
\toprule
performance & class             &          & \multicolumn{4}{c}{LR-soft} & \multicolumn{3}{c}{ideal}                              \\
            &                   &          & {strong}                    & {prd} & {weak} & {(crisp)} & {strong} & {prd} & {weak} \\
sens        & N                 & $\bar x$ & 0.541                       & 0.580 & 0.620  & 0.672     & 0.859    & 0.913 & 1.000  \\
            &                   & $s (x)$  & 0.013                       & 0.013 & 0.013  & 0.017     &          &       &        \\
            & A\textdegree II   & $\bar x$ & 0.315                       & 0.428 & 0.570  & 0.455     & 0.638    & 0.763 & 1.000  \\
            &                   & $s (x)$  & 0.007                       & 0.008 & 0.009  & 0.012     &          &       &        \\
            & A\textdegree III+ & $\bar x$ & 0.504                       & 0.554 & 0.617  & 0.620     & 0.815    & 0.876 & 1.000  \\
            &                   & $s (x)$  & 0.007                       & 0.007 & 0.008  & 0.008     &          &       &        \\
spec        & N                 & $\bar x$ & 0.799                       & 0.820 & 0.842  & 0.833     & 0.923    & 0.953 & 1.000  \\
            &                   & $s (x)$  & 0.004                       & 0.004 & 0.004  & 0.005     &          &       &        \\
            & A\textdegree II   & $\bar x$ & 0.659                       & 0.694 & 0.738  & 0.742     & 0.888    & 0.926 & 1.000  \\
            &                   & $s (x)$  & 0.006                       & 0.006 & 0.006  & 0.009     &          &       &        \\
            & A\textdegree III+ & $\bar x$ & 0.767                       & 0.802 & 0.846  & 0.851     & 0.871    & 0.914 & 1.000  \\
            &                   & $s (x)$  & 0.005                       & 0.005 & 0.005  & 0.005     &          &       &        \\
\bottomrule
\end{tabular}
}
                                                                                             \\
\subfloat[\label{tab-results-MAE} The regression-type operators.]{
\begin{tabular}{lllSSS}
\toprule
performance & class             &          & \multicolumn{3}{c}{LR-soft}                     \\
            &                   &          & {$1 - \wMAE$} & {$1 - \wRMAE$} & {$1 - \wRMSE$} \\
sens        & N                 & $\bar x$ & 0.645         & 0.404          & 0.563          \\
            &                   & $s (x)$  & 0.013         & 0.011          & 0.012          \\
            & A\textdegree II   & $\bar x$ & 0.584         & 0.355          & 0.508          \\
            &                   & $s (x)$  & 0.007         & 0.006          & 0.007          \\
            & A\textdegree III+ & $\bar x$ & 0.648         & 0.407          & 0.568          \\
            &                   & $s (x)$  & 0.006         & 0.005          & 0.006          \\
spec        & N                 & $\bar x$ & 0.829         & 0.586          & 0.712          \\
            &                   & $s (x)$  & 0.004         & 0.005          & 0.006          \\
            & A\textdegree II   & $\bar x$ & 0.734         & 0.484          & 0.664          \\
            &                   & $s (x)$  & 0.006         & 0.006          & 0.006          \\
            & A\textdegree III+ & $\bar x$ & 0.824         & 0.581          & 0.744          \\
            &                   & $s (x)$  & 0.004         & 0.005          & 0.006          \\
\bottomrule
\end{tabular}
}
\end{table}
\processdelayedfloats
\end{document}